%% file: main.tex
\documentclass[10pt,twocolumn,letterpaper]{article}

\usepackage{cvpr}              
\usepackage{multirow}
\usepackage[accsupp]{axessibility}

\input{preamble}

\usepackage{array}
\usepackage{diagbox}
\usepackage{marvosym}
\usepackage{tcolorbox}

\definecolor{cvprblue}{rgb}{0.21,0.49,0.74}
\usepackage[pagebackref,breaklinks,colorlinks,citecolor=cvprblue]{hyperref}
\newcolumntype{C}[1]{>{\centering\arraybackslash}p{#1}}
\newcolumntype{L}[1]{>{\raggedright\arraybackslash}p{#1}}

\def\paperID{4666} 
\def\confName{CVPR}
\def\confYear{2025}

\title{Marten\includegraphics[width=0.03\textwidth, keepaspectratio]{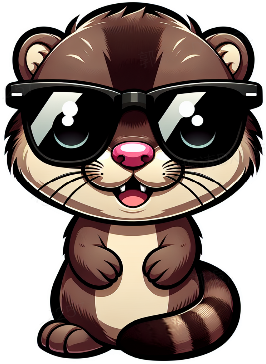}: Visual Question Answering with Mask Generation for Multi-modal Document Understanding}

\author{Zining Wang\textsuperscript{\rm 1\thanks{These authors contributed equally. $^{\textrm{\Letter}}$Corresponding Author.}}, Tongkun Guan\textsuperscript{\rm 2\footnotemark[1]}, Pei Fu\textsuperscript{\rm 1(\textrm{\Letter})},
Chen Duan\textsuperscript{\rm 1},
Qianyi Jiang\textsuperscript{\rm 1},
Zhentao Guo\textsuperscript{\rm 3},
Shan Guo\textsuperscript{\rm 1},\\
Junfeng Luo\textsuperscript{\rm 1}, 
Wei Shen\textsuperscript{\rm 2(\textrm{\Letter})}, Xiaokang Yang\textsuperscript{\rm 2}
\\
\textsuperscript{\rm 1} Meituan 
\textsuperscript{\rm 2} MoE Key Lab of Artificial Intelligence, AI Institute, Shanghai Jiao Tong University\\
\textsuperscript{\rm 3} Beijing Institute of Technology\\
{\tt\small \{wangzining03,fupei\}@meituan.com, \{gtk0615,wei.shen\}@sjtu.edu.cn}
}

\begin{document}
\maketitle
\input{sec/0_abstract}    
\input{sec/1_intro}

\input{sec/2_related_work}

\input{sec/3_method}
\input{sec/4_experiment}
\section{Acknowledgements} 
This work was supported by NSFC 62322604, NSFC 62176159 and Shanghai Municipal Science and Technology Major Project 2021SHZDZX0102.
{
    \small
    \bibliographystyle{ieeenat_fullname}
    \bibliography{main}
}

\newpage
\include{suppl}

\end{document}

%% file: preamble.tex
\usepackage[dvipsnames]{xcolor}


%% file: sec/0_abstract.tex
\begin{abstract}
Multi-modal Large Language Models (MLLMs) have introduced a novel dimension to document understanding, i.e., they endow large language models with visual comprehension capabilities; however, how to design a suitable image-text pre-training task for bridging the visual and language modality in document-level MLLMs remains underexplored.
In this study, we introduce a novel visual-language alignment method that casts the key issue as a \textbf{V}isual \textbf{Q}uestion \textbf{A}nswering with \textbf{Mask} generation (\textbf{VQAMask}) task, optimizing two tasks simultaneously: VQA-based text parsing and mask generation. The former allows the model to implicitly align images and text at the semantic level. The latter introduces an additional mask generator (discarded during inference) to explicitly ensure alignment between visual texts within images and their corresponding image regions at a spatially-aware level. Together, they can prevent model hallucinations when parsing visual text and effectively promote spatially-aware feature representation learning. To support the proposed VQAMask task, we construct a comprehensive image-mask generation pipeline and provide a large-scale dataset with 6M data (MTMask6M). Subsequently, we demonstrate that introducing the proposed mask generation task yields competitive document-level understanding performance. Leveraging the proposed VQAMask, we introduce Marten, a training-efficient MLLM tailored for document-level understanding. 
Extensive experiments show that our Marten consistently achieves significant improvements among 8B-MLLMs in document-centric tasks. Code and datasets are available at \url{https://github.com/PriNing/Marten}.
\end{abstract}

%% file: sec/1_intro.tex
\vspace{-1.5em}
\section{Introduction}
\vspace{-0.5em}
\label{sec:intro}
Large Language Models (LLMs) have shown a comprehensive generalization ability across a wide range of language-related tasks~\cite{bai2023qwen,achiam2023gpt}. These successful experiences have inspired researchers to explore Multi-modal Large Language Models (MLLMs) in the context of Visual Question Answering (VQA), \emph{i.e.,} empower the LLMs with visual comprehension capabilities. 
However, a significant challenge arises in understanding text within document images, possibly due to high resolution, densely packed, small visual texts, and diverse image forms.
\begin{figure}[t]
  \centering
  \includegraphics[width=0.47\textwidth]{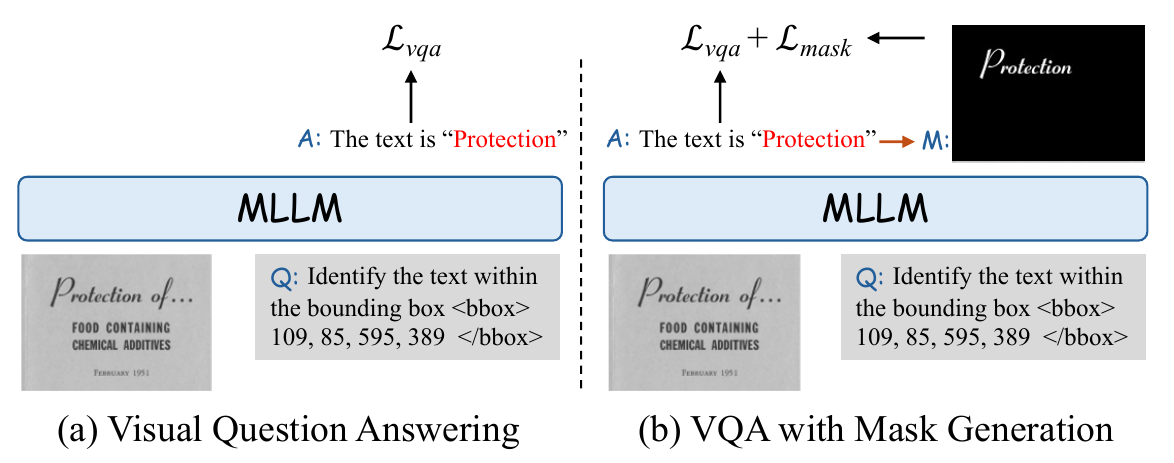}
  \caption{Different pre-training paradigms of MLLMs for document understanding: (a) \emph{Visual Question Answering} (VQA) paradigm that implicitly aligns visual and language modality at the semantic level; (b) our proposed \emph{Visual Question Answering with Mask generation} (VQAMask) paradigm. Building on VQA, we introduce an additional Mask Generator during training to explicitly align visual texts and their corresponding image regions at a spatially-aware level. During the inference stage, the mask generator is discarded.}
  \label{fig1}
  \vspace{-1em}
\end{figure}

To enhance visual comprehension, several studies~\cite{LayTextLLM,MOAI,Cream,Instructdoc,Doclayllm,internvl,minigpt,MiNi-Monkey,Monkey,Mplug-Docowl2,UReader} have focused on designing pre-training tasks specifically tailored for document images to achieve visual-language alignment. These tasks include full-text recognition or transcription, text spotting, and visual text grounding, \emph{etc}, following various prompts. For instance,  KOSMOS-2.5 \cite{lv2023kosmos} proposes a visual text grounding task, which inputs the texts within images and produces the corresponding bounding boxes. Vary \cite{vary} and mPLUG-DocOWL \cite{mplug-docowl1.5} introduce the learning of struct-aware document parsing, table parsing, chart parsing and natural image parsing for different image forms to enhance fine-grained textual perception. 

Although existing methods demonstrate promising capabilities, we argue that these training tasks predominantly emphasize \emph{semantic alignment} and only implicitly capture the spatial location of text within document images. However, \emph{spatial alignment} is also a crucial factor for accurately interpreting document images.
Without spatially-aware supervision, the outputs may disproportionately rely on the powerful semantic context capabilities of large language models (LLMs) rather than optimizing image features from visual encoders, potentially leading to model hallucinations.




To address this issue, we propose a novel vision-language alignment method for visual document understanding, \textbf{V}isual \textbf{Q}uestion \textbf{A}nswering with \textbf{Mask} generation (\textbf{VQAMask}), to explicitly facilitate spatially-aware feature representation learning. As illustrated in Figure \ref{fig1} (b), visual tokens and language tokens are input into the LLM to jointly optimize two tasks: VQA-based text parsing and mask generation. For the task of VQA-based text parsing, the model predicts the corresponding answer, following different OCR-related prompts. This task can facilitate the model to align images and text at the semantic level. For the task of mask generation, we introduce an additional \textbf{M}ask \textbf{G}enerator \textbf{M}odule (\textbf{MGM}) to explicitly align images and text at the spatially-aware level. 
Specifically, in the intermediate layer of the LLM, we take the cross-attention interaction between the part of the visual modality (query) and the part of the language modality (key) to obtain attention maps. These attention maps, followed by several deconvolution layers, are restored to the original image resolution. Subsequently, we constrain them to ensure spatial alignment between visual texts within images and their corresponding image regions, under the groundtruth mask supervision constructed by our established mask acquisition pipeline.
Additionally, it is important to note that this mask generation task is discarded during the inference stage, and it does not add any additional cost to the inference process. Experiments demonstrate the proposed VQAMask works well in various visual encoders and language models. 

Utilizing the proposed VQAMask, we introduce a training-efficient MLLM, Marten, which consistently achieves significant improvements among 8B-MLLMs in document-centric tasks.
Our contributions are as follows:
\begin{itemize}
    \item[1)] We introduce a novel Visual Question Answering with Mask generation (VQAMask) task to facilitate spatially-aware and semantic-aware feature representation learning for visual language alignment.
    \item[2)] We establish a mask acquisition pipeline to generate mask labels without manual annotation, and provide a large-scale dataset (MTMask6M) with 6M image-mask pairs. 
    \item[3)] Extensive experiments demonstrate the effectiveness of the VQAMask task and outperform the previous state-of-the-art method by 0.4\%, 0.4\%, 6.2\%, 1.8\%, 6.2\%, 4.0\%, 1.5\%, and 10.1\% on DocVQA, InfoVQA, DeepForm, KLC, WTQ, TabFact, FUNSD, and SROIE datasets.
    
\end{itemize}

%% file: sec/2_related_work.tex
\section{Related Work}
\label{sec:formatting}

\subsection{Multi-modal Document Understanding}
Multi-modal Document Understanding aims to extract meaningful information from text images of various types, such as charts, tables, documents, and other scene texts, through a question-driven image-to-sequence task.
Some early studies~\cite{yang2021tap} have explored end-to-end solutions within a specialist model, which may not provide broad robustness and generality for various scenarios.
The recent emergence of Multi-modal Large Language Models (MLLMs) has introduced a novel dimension to the field by linking visual image tokens and language tokens in a sequence-to-sequence format, thereby facilitating task unification. This structure seamlessly integrates computer vision with natural language processing, allowing MLLMs to significantly enhance text reading capabilities, supported by large-scale data and GPU resources. These methods can be roughly categorized into two types: OCR-dependent MLLMs~\cite{LayTextLLM,MOAI,Cream,Instructdoc,Doclayllm} and OCR-free MLLMs~\cite{internvl,minigpt,MiNi-Monkey,Monkey,Mplug-Docowl2,UReader}.

\noindent \textbf{OCR-dependent MLLMs} enhance document understanding by integrating text, layout, and other data extracted from external OCR tools~\cite{PP-OCRv3} into large language models. LayTextLLM~\cite{LayTextLLM} and DocLayLLM~\cite{Doclayllm} both utilize an external OCR engine to extract layout and text, integrating them into a LLM for document understanding. However, this integration complicates the workflow and leads to an excess of auxiliary tokens, particularly in images with dense texts.

\noindent \textbf{OCR-free MLLMs} perform the multi-modal document understanding task by directly producing question-driven outputs in an end-to-end manner. These methods typically focus on high-resolution image processing~\cite{Monkey,UReader,mplug-docowl1.5,docpedia}, efficient token compression~\cite{zhang2024token,Mplug-Docowl2,yu2024texthawk2}, and refined attention mechanisms~\cite{MiNi-Monkey,VisualCoT}. In the study, we focus on exploring suitable pre-training tasks, tailored for document images.

\begin{figure*}[t]
  \centering
  \includegraphics[width=0.97\textwidth]{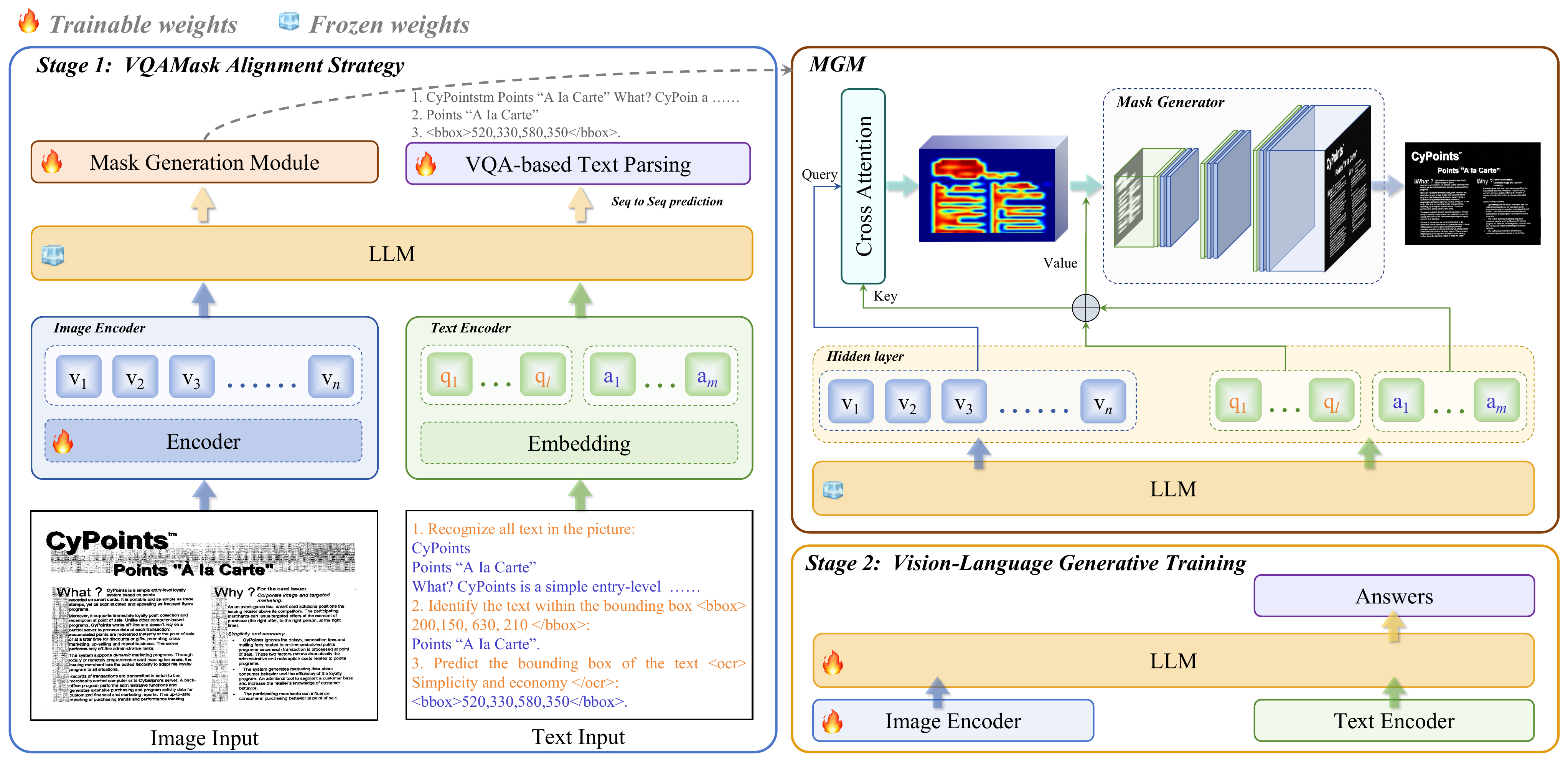}
  \vspace{-1.0em}
  \caption{Overview of our proposed Marten architecture. The training of the model is divided into two stages: \textbf{1) VQAMask Alignment Training}: the proposed vision-language alignment method, VQAMask, includes two pre-training tasks: VQA-based text parsing and mask generation. By integrating these two tasks, VQAMask not only effectively enables the Marten model to implicitly learn the visual text within images at the semantic level but also explicitly aligns images and text at the spatially-aware level; \textbf{2) Vision-Language Generative Training}: In the stage, we discard the mask generation task. A wide range of high-quality instruction data is collected to conduct VQA tasks for general document-level understanding.}
  \vspace{-1.0em}
  \label{fig2}
\end{figure*}

\subsection{Vision Language Pre-training}
Inspired by recent advancements~\cite{CLIP,DINO,SigLIP,SIGA,SAM} in pre-training techniques, the integration of image and text multi-modal information into OCR-related tasks has gained increasing attention. Using cross-modal visual-language priors, early works focused on endowing visual foundation models with semantic knowledge~\cite{FastTCM,TCM,SKTM,VLPT,oCLIP,duan2024odm,guan2025bridging,CCD} for applications such as text spotting, detection~\cite{guan2024bridging,guan2022industrial}, recognition~\cite{guan2023self,guan2025ccdplus,guan2024posformer,lyu2022maskocr}, removal, and super-resolution. As MLLMs rapidly develop, researchers are further capitalizing on these visual-language priors to bridge visual and language modalities through diverse pre-training tasks~\cite{UReader,Cream,Doclayllm,DocLLM,docpedia,LayTextLLM,internvl,liu2024textmonkey,minigpt,mplug-docowl1.5}. For instance, UReader~\cite{UReader} introduces the Read Full Text (RFT) task in VQA form for enhancing document-level understanding. Park \emph{et al.}~\cite{parkhierarchical} propose two new pretext tasks: Reading Partial Text (RPT) and Predicting Text Position (PTP). Similarly, KOSMOS-2.5~\cite{lv2023kosmos} designs a Visual Text Grounding (VTG) task, which inputs the texts within images and produces the corresponding bounding boxes. mPLUG-DocOWL~\cite{mplug-docowl1.5} integrates multiple tasks to conduct the struct-aware parsing in documents, tables, charts, and natural scenes. However, these question-driven image-to-sequence tasks predominantly emphasize \emph{semantic alignment}, and may rely on the powerful semantic context capabilities of LLMs when responding. Following these VQA forms, we further introduce an additional mask generation pre-text task (VQAMask) to explicitly facilitate spatially-aware visual-language alignment.

%% file: sec/3_method.tex
\section{Methodology}
In this section, we first review the representative MLLM method that connects the visual modality and language modality into LLM to generate responses. Building on this foundation, we present our proposed pre-training method, Visual Question Answering with Mask generation (VQAMask), designed specifically for Multi-modal Document Understanding.

\noindent \textbf{Preliminary.}
Typically, Multi-modal Large Language Models (MLLMs) include a visual foundation model (VFM), a modality connector, and a large language model (LLM). Initially, following the prevalent multi-scale adaptive cropping strategy, the input high-resolution image $\mathbf{X}\in \mathbb{R}^{H \times W}$ is first cropped into several non-overlapping sub-images. $H$ and $W$ represent the image height and width. These sub-images are then processed by the visual foundation model to obtain image patches, concretely represented by $[\mathbf{x}_{1},...,\mathbf{x}_{n}]$, along with their corresponding visual embeddings $\mathbf{V} = [\mathbf{v}_{1},...,\mathbf{v}_{n}]$. Here, $n$ denotes the number of image patches. For language input, the question and the answer (option for training) is tokenized using the BPE tokenizer, resulting in $l$ question tokens embedded as $\mathbf{Q} = [\mathbf{q}_{1},...,\mathbf{q}_{l}]$ and $m$ answer tokens embedded as $\mathbf{A} = [\mathbf{a}_{1},...,\mathbf{a}_{m}]$. Subsequently, the modality connector acts a bridge between the visual embeddings and language (question and answer) embeddings. Finally, the visual embeddings $\mathbf{V}$ and language embeddings $\mathbf{Q}$ and $\mathbf{A}$ are fed into the LLM to generate more precise and comprehensive answers. Specifically, the LLM process can be described as:
{\setlength\abovedisplayskip{2pt}
\setlength\belowdisplayskip{2pt}
\begin{align}
\mathbf{V}^{k+1},\mathbf{Q}^{k+1},\mathbf{A}^{k+1} = 
\mathrm{Layer_{LLM}}(&[\mathbf{V}^{k},\mathbf{Q}^{k},\mathbf{A}^{k},\mathbf{E}])
\label{eq1}
\end{align}
}

\noindent where $k \in \{0,1,...,K\}$ and $\mathbf{V}^{k+1}$, $\mathbf{Q}^{k+1}$, and $\mathbf{A}^{k+1}$ refers to the outputs of $k$-th layer of LLM. $\mathbf{E}$ refers to the attention mask, which is usually a lower triangular matrix used to prevent attention to certain positions. Note that we omit the superscript for $k=0$, because these vectors are the initial value fed into the LLM. 
During inference, the input answer tokens $\mathbf{A}$ are replaced as the previous predicted tokens. 

Notably, some works introduce the pixel shuffle operation~\cite{internvl} to reduce the number of visual tokens, a strategy also adopted in our proposed method. However, the global operation changes the original spatial structure of these visual tokens. Differently, inspired by the Swin Transformer~\cite{liu2021swin}, we conduct pixel shuffle in each local window ($4 \times 4$ by default) to adapt to our subsequent VQAMask.

\vspace{-0.5em}
\subsection{VQA with Mask Generation}
\vspace{-0.5em}

\begin{figure*}
    \centering
    \includegraphics[width=1.\textwidth]{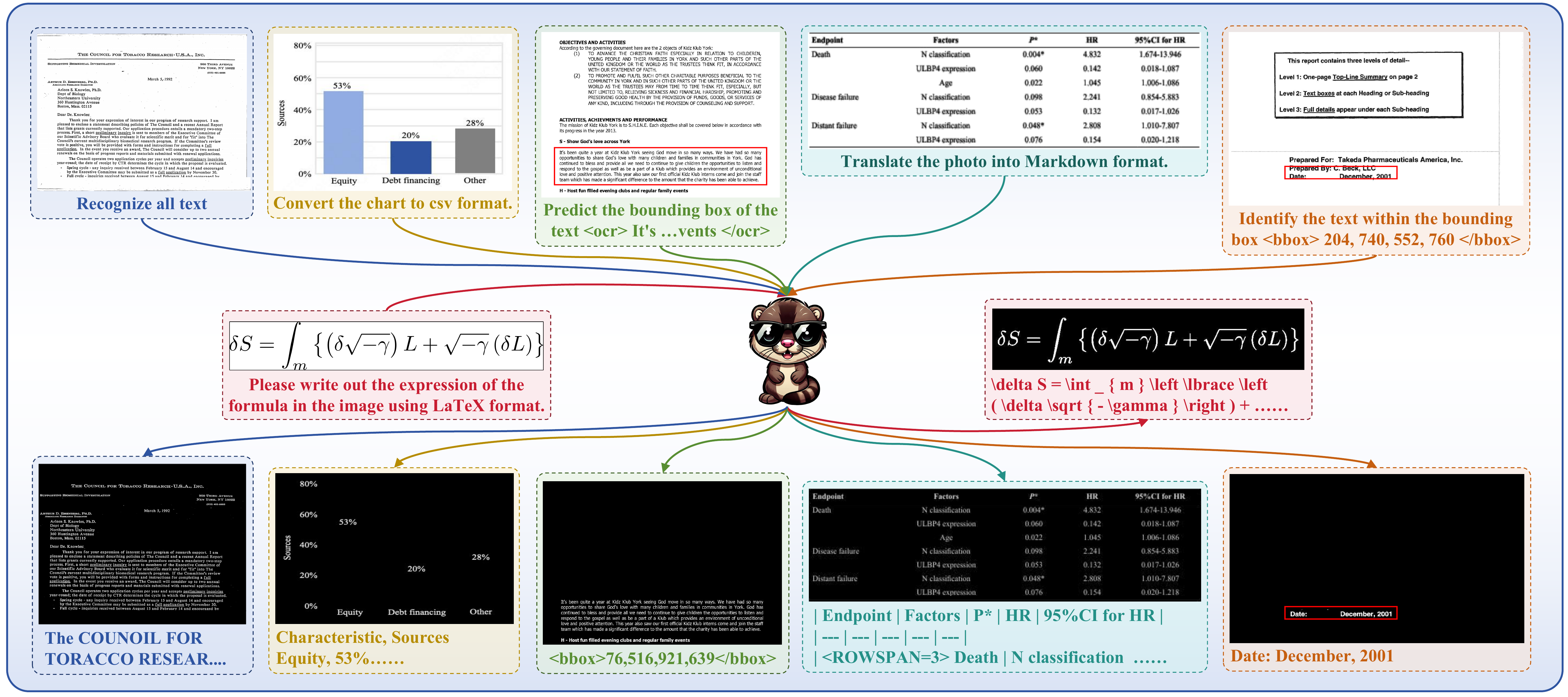}
    \caption{Illustration of the VQAMask alignment training for document parsing question answering. We introduced a total of six tasks, which can be broadly categorized into 1) Read Full Text, Reading Partial Text within Localization, and Visual Text Grounding; 2) Transcription involves converting formulas into LaTeX, tables into markdown or LaTeX, and charts into CSV and markdown formats.}
    \label{fig3}
    \vspace{-1.5em}
\end{figure*}

To bridge the gap between visual and language modality for multi-modal document understanding, previous MLLMs have formulated various pre-training tasks, including text transcription and visual text grounding, \emph{i.e.}, given customized task prompts, these methods generate prompt-related text responses. The pre-training paradigm lacks spatially-aware supervision, which may result in model hallucinations. 
To address this, we introduce a novel pre-training method, Visual Question Answering with Mask generation (VQAMask). This method incorporates an additional mask generation task to ensure spatial alignment between visual texts within images and their corresponding image regions, as illustrated in Figure \ref{fig2}.
Specifically, the proposed VQAMask includes two tasks as follows:

\noindent \textbf{VQA-based Text Parsing.} Following existing works~\cite{mplug-docowl1.5,Mplug-Docowl2,liu2024textmonkey,vary}, we introduce the text parsing task to implicitly align images and text at the semantic level. The specific task prompts are presented in Figure \ref{fig3}. The outputs of the last layer of LLM are utilized to predict these answers, and the optimization loss is formulated as follows:
{\setlength\abovedisplayskip{2pt}
\setlength\belowdisplayskip{2pt}
\begin{align}
        \mathcal{L}_{vqa} = \mathbf{A} \log p(\mathbf{A}^{K+1}|\mathbf{V},\mathbf{Q}) 
\end{align}
}

\noindent \textbf{Mask Generation.} In the subsection, we integrate a mask generation module (MGM) into the hidden layers of the LLM to explicitly enhance vision-language alignment at a spatial-aware level.
Specifically, we first feed the hidden states ($\mathbf{V}^{k}$, $\mathbf{Q}^{k}$, and $\mathbf{A}^{k}$) of the selected layer $k-1$ into a four-layer transformer module, with each layer including two sub-layers: a multi-head cross-attention mechanism, and a positionwise fully connected feed-forward network. The specific implementation is as follows:
{\setlength\abovedisplayskip{2pt}
\setlength\belowdisplayskip{2pt}
\[
\left\{
\begin{aligned}
&\mathbf{H}^{k} = [\mathbf{Q}^{k},\mathbf{A}^{k}]\\
&\mathbf{Attn}= \sigma\left(\frac{\mathbf{V}^{k}\mathbf{W}_{query}\cdot(\mathbf{H}^{k}\mathbf{W}_{key})^\intercal }{\sqrt{d}}\right)\mathbf{H}^{k}\mathbf{W}_{value}, \\
&\mathbf{V}_{attn} = \max(0,\mathbf{Attn}\cdot \mathbf{W}_1+\mathbf{b}_1)\mathbf{W}_2+\mathbf{b}_2,
\end{aligned}
\right.
\]
}

\noindent where $[\cdot]$ denotes the concatenation operation and $\sigma (\cdot)$ refers to the softmax activate function.
The projections are parameter matrices $\mathbf{W}_1, \mathbf{W}_2,\mathbf{W}_{query}, \mathbf{W}_{key}, \mathbf{W}_{value}$ and $d$ denotes the dimension.

By fully interacting with the answer tokens, the visual tokens corresponding to the visual text regions are highlighted. Subsequently, these one-dimensional visual tokens are re-organized into two-dimensional image space. Followed by several transposed convolutions $\phi$, we then restore these visual tokens to the resolution of the input image. The specific process is as follows:
{\setlength\abovedisplayskip{2pt}
\setlength\belowdisplayskip{2pt}
\begin{align}
\Tilde{\mathbf{M}}=\phi(\mathbf{V}_{attn})
\end{align}
}

\noindent where $\Tilde{\mathbf{M}}$ refers to the final predicted mask.
Finally, a Dice loss~\cite{milletari2016vdiceloss} and Cross-Entropy loss are employed to optimize the segmentation network:
{\setlength\abovedisplayskip{2pt}
\setlength\belowdisplayskip{2pt}
\begin{align}
    \mathcal{L}_{mask} = l_{\rm DICE}(\Tilde{\mathbf{M}},\mathbf{M}) + l_{\rm CE}(\Tilde{\mathbf{M}}, \mathbf{M})
\end{align}
}

\noindent where $\mathbf{M}$ denotes the groundtruth mask of the input image, which will be introduced in Sec. \ref{sec3.2}.

\subsection{Mask Acquisition Pipeline}
\label{sec3.2}


We note that in document scenarios, the boundary between text and background is typically distinct, allowing for easy separation of text from the entire image using a threshold. Previous research, such as CCD~\cite{CCD}, has explored and confirmed this observation. Inspired by CCD~\cite{CCD}, we propose a clustering-based binarization method for foreground construction, comprising three stages: preparation, clustering, and generation. The specific process is as follows:

\noindent \textbf{Preparation.} We utilize PaddleOCR~\cite{PP-OCRv3} to detect all visual text regions within an image and obtain corresponding cropped text instance images based on the bounding boxes.

\noindent \textbf{Clustering.} For each cropped text instance image, we employ a simple yet effective clustering model (K-means) to classify image pixels into two clusters. Given that visual text tends to be concentrated in the center region of an image, we calculate the distance of the pixels in each cluster from the center position of the cropped image. The cluster with pixels closer to the center is identified as the foreground (with a pixel value of 1), while the other is identified as the background (with a pixel value of 0).
Subsequently, a secondary calibration is conducted to verify the correctness of the obtained foreground cluster. Specifically, we compare the average pixel value of the edge regions of the cropped text instance image with the overall average pixel value. If the former is higher, a 0-1 inversion is implemented.

\noindent \textbf{Generation.} These foreground masks from all cropped text instance images are reassembled according to their original coordinates to obtain a complete mask image.

\subsection{Training Strategy}

As shown in Figure \ref{fig2}, we divide the training process into two stages: our proposed VQAMask vision-language alignment training and vision-language generative training. 

\noindent \textbf{Stage 1: VQAMask Alignment Training.} 
Currently, most MLLMs for document understanding implement image-text alignment to bridge the visual foundation model with the LLM as the first training stage task. Although such alignment methods endow the MLLMs with basic text recognition capabilities, they lack spatial awareness of the visual text within images. As a result, they struggle to accurately locate complex text within text-rich images and understand the structural information of documents.
To enhance the spatial awareness of visual text in documents, we propose a VQAMask vision-language alignment training to bridge the visual foundation model with the language model. 

Regarding data usage, as displayed in Table \ref{tab:stage1}, we utilized the pipeline proposed in Section \ref{sec3.2} to construct 6 million samples, referred to as MTMask6M, including DocStruct4M~\cite{mplug-docowl1.5}, IIT-CDIP~\cite{iit} and DocGenome~\cite{xia2024docgenome}, Scene Text Datasets~\cite{karatzas2013icdar,karatzas2015icdar,gupta2016synthetic,singh2021textocr}. Each sample includes the image, question, answer, and the corresponding mask. DocStruct4M is provided by DocOwl 1.5~\cite{mplug-docowl1.5}, which includes five categories: natural images, documents (CCPdf~\cite{turski2023ccpdf}, DUE~\cite{borchmann2021due}, RVL-CDIP~\cite{harley2015evaluation}), tables (TURL~\cite{deng2022turl}, PubTabNet~\cite{zhong2020image}), charts (ChartQA~\cite{masry2022chartqa}, FigureQA~\cite{kahou2017figureqa}, DVQA~\cite{kafle2018dvqa}, PlotQA~\cite{methani2020plotqa}), and web pages (VisualMRC~\cite{tanaka2021visualmrc}). 
DocGenome~\cite{xia2024docgenome} includes three types of scenarios: documents, formulas, and tables. We name them DocGenome-D, DocGenome-F, and DocGenome-T, respectively. Additionally, we also introduce ICDAR13~\cite{karatzas2013icdar}, ICDAR15~\cite{karatzas2015icdar}, SynthText~\cite{gupta2016synthetic}, TextOCR~\cite{singh2021textocr}, and OpenVINO~\cite{krylov2021open} as the scene text datasets. 



For VQA-based text parsing tasks, we construct six types of QA pairs, as illustrated in Figure \ref{fig3}. These include reading full text, visual text recognition with coordinates, visual text grounding, and markdown, LaTeX and CSV format transcription. 

During pre-training, the weights of vision foundation model, MLP, and MGM are updated, while the LLM remains frozen. The goal is to preserve the inherent semantic context capability of the LLM while specifically enhancing the overall MLLM's spatial awareness for visual texts within images.

\noindent \textbf{Stage 2: Vision-Language Generative Training.} 
In this stage, we collect existing VQA datasets related to document understanding scenarios, as shown in Table \ref{tab:stage2}. A generative training strategy is then employed to enhance general document-level comprehension. Specifically, our visual foundation model and MLP inherit the weights from stage 1. The proposed MGM is discarded in this stage. Additionally, we unfreeze the weights of LLM
and all parameters are updated to conduct supervised fine-tuning (SFT), which includes full data training and high-quality data fine-tuning. First, all data are included in the full data training. Then we collect a batch of high-quality instruction data (one-tenth) from the full dataset to fine-tune the model. The detailed data usage is summarized in Table \ref{tab:stage2}.
Through the combined training of these two phases, our model is able to extract more powerful visual representations and exhibits significantly enhanced document understanding capabilities. \label{stage_2}

\begin{table}[t]
    \centering
    \caption{Details of MTMask6M used in VQAMask Alignment Training (Stage 1). 
    ``Text R/G" refers to visual text recognition and visual text grounding, with data sourced from the Multi-Grained Text Localization section of DocStruct4M.} 
    \scalebox{0.9}{
    \begin{tabular}{c|c|l}
    \hline
    Task & Samples & Datasets \\
    \hline
        \multirow{3}{*}{\begin{tabular}[c]{@{}c@{}}Document\\ parsing\end{tabular}} & \multirow{3}{*}{3361.3k} & IIT-CDIP~\cite{iit}, CCPdf~\cite{turski2023ccpdf}\\
        & &DUE~\cite{borchmann2021due}, VisualMRC~\cite{tanaka2021visualmrc} \\
        & & RVL-CDIP~\cite{harley2015evaluation}, DocGenome-D~\cite{xia2024docgenome} \\
    \hline
        \multirow{2}{*}{\begin{tabular}[c]{@{}c@{}}Table\\ parsing\end{tabular}} & \multirow{2}{*}{600k} & TURL~\cite{deng2022turl}, PubTabNet~\cite{zhong2020image} \\
        & & DocGenome-T~\cite{xia2024docgenome}\\
    \hline
        \multirow{2}{*}{\begin{tabular}[c]{@{}c@{}}Chart\\ parsing\end{tabular}} & \multirow{2}{*}{475.1k} & ChartQA~\cite{masry2022chartqa}, FigureQA~\cite{kahou2017figureqa}\\
        & & DVQA~\cite{kafle2018dvqa}, PlotQA~\cite{methani2020plotqa}\\
    \hline
        \begin{tabular}[c]{@{}c@{}}Formula\\ parsing\end{tabular} & 200k & DocGenome-F~\cite{xia2024docgenome} \\
    \hline
        \multirow{3}{*}{\begin{tabular}[c]{@{}c@{}}Scene text\\ parsing\end{tabular}} & \multirow{3}{*}{395.6k} & ICDAR13~\cite{karatzas2013icdar}, ICDAR15~\cite{karatzas2015icdar} \\
        & & SynthText~\cite{gupta2016synthetic}, Textocr~\cite{singh2021textocr} \\
        & & OpenVINO~\cite{krylov2021open} \\
    \hline 
    \multirow{1}{*}{Text R/G} & \multirow{1}{*}{1000k} & DocStruct4M-subset~\cite{mplug-docowl1.5} \\
    \hline
    Total & \multicolumn{2}{c}{6032k} \\
    \hline
    \end{tabular}
    }
    \label{tab:stage1}

    \vspace{1em}
    
    \caption{Details of the training datasets used in Vision-Language Generative Training (Stage 2). $\dag$ denotes the selected high-quality instruction data, utilized for supervised fine-tuning again.}
    \scalebox{0.9}{
    \begin{tabular}{c|c|l}
    \hline
    Task & Samples & Datasets \\
    \hline
    \multirow{3}{*}{Document VQA} & \multirow{3}{*}{2301.5k} & DocVQA$\dag$~\cite{mathew2021docvqa}, InfoVQA$\dag$~\cite{mathew2022infographicvqa}\\
    & &DeepForm$\dag$~\cite{svetlichnaya2020deepform}, KLC$\dag$~\cite{stanislawek2021kleister}\\
    & & DocMatix~\cite{laurençon2024building} \\
    \hline
    \multirow{2}{*}{Table VQA} & \multirow{2}{*}{107.6k} & TableFact$\dag$~\cite{chen2019tabfact}, WTQ$\dag$~\cite{pasupat2015compositionalwtq}\\
    & &TableBench~\cite{wu2024tablebench}\\
    \hline
    \multirow{2}{*}{Chart VQA} & \multirow{2}{*}{318.1k} & ChartQA$\dag$~\cite{masry2022chartqa}, FigureQA~\cite{kahou2017figureqa}\\
    & &DVQA$\dag$~\cite{kafle2018dvqa}\\
    \hline
    \multirow{2}{*}{Formula VQA} & \multirow{2}{*}{274.5k} & UniMER~\cite{wang2024unimernet}, \\
    & & CROHME$\dag$~\cite{mahdavi2019icdar,mouchere2014icfhr, mouchere2016icfhr2016}\\
    \hline
    \multirow{3}{*}{Sence Text VQA} & \multirow{3}{*}{289.3k} & TextVQA$\dag$~\cite{singh2019towardstextvqa}, ST-VQA$\dag$~\cite{biten2019scene}\\
    & & OCR-VQA~\cite{mishra2019ocr}, IAM~\cite{marti2002iam}$\dag$\\
    & & EST-VQA~\cite{wang2020general} \\
    \hline
    \multirow{1}{*}{KIE} & \multirow{1}{*}{6.2k} & FUNSD$\dag$~\cite{jaume2019funsd}, SROIE$\dag$~\cite{huang2019icdar2019sroie}\\
    \hline
    Total & \multicolumn{2}{c}{3297.2k} \\
    \hline
    \end{tabular}
    }
    \label{tab:stage2}
    \vspace{-1.em}
\end{table}

%% file: sec/4_experiment.tex
\section{Experiments}
\begin{table*}[t]
    \centering
    \caption{Comparison with OCR-free methods on various types of text-rich image understanding tasks. All evaluation benchmarks use the officially designated metrics. ``size" refers to the number of parameters in the model, and ``Val" refers to the validation set.}
    \vspace{-0.5em}
    \scalebox{0.73}{
    \begin{tabular}{c|c|c|cccc|c|c|cc|ccc}
        \toprule 
        Model & size &Venue & DocVQA & InfoVQA & DeepForm & KLC & ChartQA & TextVQA\textsubscript{Val} & WTQ & TabFact & FUNSD & SROIE  & POIE \\
        \midrule
        DocPeida~\cite{feng2023docpedia} & 7.1B &arxiv'23 & 47.1 & 15.2 & - & - & 46.9 & 60.2 & - & - & 29.9 & 21.4 & 39.9   \\
        DocOwl~\cite{ye2023mplug} & 7.3B &arxiv'23 & 62.2 & 38.2 & 42.6 & 30.3 & 57.4 & 52.6 & 26.9 & 67.6 & 0.5 & 1.7 & 2.5   \\
        LLaVA1.5~\cite{liu2024visual} & 7.3B & NeurIPS'23 & - & - & - & - & 9.3 & - & - & - & 0.2 & 1.7 & 2.5 \\
        UReader~\cite{UReader} & 7.1B &EMNLP'23 & 65.4 & 42.2 & 49.5 & 32.8 & 59.3 & 57.6 & 29.4 & 67.6 & - & - & -   \\ 
        CHOPINLLM~\cite{fan2024pre} & 7B &arxiv'24 & - & - & - & - & 69.98 & - & - & - & - & - & -   \\ 
        TextHawk~\cite{yu2024texthawk} & 7.4B &arxiv'24 & 76.4 & 50.6 & - & - & 66.6 & - & 34.7 & 71.1 & - & - & -   \\ 
        DocKylin~\cite{zhang2024dockylin} & 7.1B &arxiv'24 & 77.3 & 46.6 & - & - & 66.8 & - & 32.4 & - & - & - & -   \\ 
        MM1.5~\cite{zhang2024mm1} & 7.3B &arxiv'24 & 88.1 & 59.5 & - & - & 78.6 & \underline{76.8} & 46.0 & 75.9 & - & - & -   \\ 
        Mini-Monkey~\cite{MiNi-Monkey} & 2B &arxiv'24 & 87.4 & 60.1 & - & - & 76.5 & 75.7 & - & - & \underline{42.9} & \underline{70.3} & \textbf{69.9}  \\
        DocOwl-1.5~\cite{mplug-docowl1.5} & 8.1B &EMNLP'24 & 81.6 & 50.4 & 68.8 & 37.9 & 70.5 & 68.8 & 39.8 & \underline{80.4} & - & - & -   \\
        DocOwl-1.5-Chat~\cite{mplug-docowl1.5} & 8.1B &EMNLP'24 & 82.2 & 50.7 & \underline{68.8} & \underline{38.7} & 70.2 & 68.6 & 40.6 & 80.2 & - & - & -  \\
        CogAgent~\cite{hong2024cogagent} & 17.3B &CVPR'24 & 81.6 & 44.5 & - & - & 68.4 & 76.1 & - & - & - & - & -   \\
        Monkey~\cite{Monkey} & 9.8B &CVPR'24 & 66.5 & 36.1 & 40.6 & - & 65.1 & 67.6 & 25.3 & - & - & - & -   \\
        TextMonkey~\cite{liu2024textmonkey} & 7.7B &arxiv'24 & 73.0 & 28.6 & - & - & 66.9 & 65.6 & - & - & 32.3 & 47.0 & 27.9 \\
        HRVDA~\cite{liu2024hrvda} & 7.1B & CVPR'24 & 72.1 & 43.5 & 63.2 & 37.5 & 67.6 & 73.3 & 31.2 & 72.3 & - & - & - \\
        InternVL2~\cite{internvl} & 8.1B & CVPR'24 & \underline{91.6} & \underline{74.8} & -  & - & \textbf{83.3} & 
        \textbf{77.4} & - & - & - & - & -    \\
        Park et al.~\cite{parkhierarchical} & 7.2B & NeurIPS'24 & 72.7 & 45.9 & 53.0 & 36.7 & 63.3 & 59.2 & 34.5 & 68.2 & - & - & -   \\
        MOAI~\cite{MOAI} & 7B &ECCV'24 & - & - & - & - & - & 67.8 & - & - & - & - & -   \\ 
        Vary~\cite{vary} & 7.4B & ECCV'24 & 76.3 & - & - & - & 66.1 & - & - & - & - & - & - \\
        TextHawk2~\cite{yu2024texthawk2} & 7.4B &arxiv'24 & 89.6 & 67.8 & - & - & 81.4 & 75.1 & \underline{46.2} & 78.1 & - & - & -   \\ 
        PDF-WuKong~\cite{xie2024wukong} & 8.5B &arxiv'24 & 76.9 & - & - & - & - & - & - & - & - & - & -   \\
        Zhang et al.~\cite{zhang2024token} & 8.1B & arxiv'24 & 78.3 & 50.2 & 65.7 & 35.9 & 68.9 & 66.6 & 38.6 & 79.3 & - & - & -  \\
        \midrule
        Marten & 8.1B & - & \textbf{92.0} & \textbf{75.2} & \textbf{75.1} & \textbf{39.5} & \underline{81.7} & 74.4 & \textbf{52.4} & \textbf{84.4} & \textbf{44.4} & \textbf{80.4} & \underline{69.5}   \\ 
        \bottomrule
    \end{tabular}
    }
    
    \label{tab:res1}
\end{table*}

\begin{table*}[t]
    \centering
    \caption{Comparison of Marten with existing OCR-free multimodal large language models on OCRBench.}
    \vspace{-0.5em}
    \scalebox{0.7}{
    \begin{tabular}{c|ccccccccc}
        \hline
        \diagbox[]{dataset}{model} & Monkey~\cite{Monkey} & TextMonkey~\cite{liu2024textmonkey} & DocOwl-1.5~\cite{mplug-docowl1.5} & MM1.5~\cite{zhang2024mm1} & TextHawk2~\cite{yu2024texthawk2} & GLM-4v~\cite{hong2024cogvlm2} & InternVL2~\cite{internvl} & MiniMonkey~\cite{MiNi-Monkey} & Marten(ours) \\
        \hline
        OCRBench & 514 & 561 & 599 & 635 & 784 & 786 & 794 & 802 & \textbf{820} \\
        \hline
    \end{tabular}}
    \label{tab:ocrbench}
\end{table*}

\subsection{Implementation Details}

\noindent \textbf{Stage 1.} 
In practical implementation, Marten selects InternViT-300M~\cite{chen2024far} as the visual foundation model, and InternLM2, a 7B large language model~\cite{cai2024internlm2}, as the language decoder. 
We employ a dynamic image-slicing strategy in which each image is cropped into a maximum of six sub-images based on the aspect ratio and resolution, with a fixed resolution of $448\times448$ for each sub-image. Subsequently, we employ the Pixel Shuffle module, compatible with VQAM, to reduce the number of tokens to 256. 
We perform one epoch on MTMask6M in Table \ref{tab:stage1}. The learning rate for the MGM module is set to 2e-4, while for other parameters, it is set to 2e-5. The batch size on each GPU is 64, and the training is conducted on 24 GPUs for two days.


\noindent \textbf{Stage 2.} 
The dataset of Table \ref{tab:stage2} is used in the stage. The learning rate and batch size are 2e-5 and 64, respectively. The training phase is conducted on 24 H800 GPUs over 56 hours. More details are introduced in Section \ref{stage_2}.


\subsection{Results}

\noindent \textbf{Text-rich Result.} We compared Marten with OCR-free multimodal large language models on 11 text-rich image benchmarks, which cover documents (DocVQA~\cite{mathew2021docvqa}, InfoVQA~\cite{mathew2022infographicvqa}, DeepForm~\cite{svetlichnaya2020deepform}, KLC~\cite{stanislawek2021kleister}), tables (WTQ~\cite{pasupat2015compositionalwtq}, TabFact~\cite{chen2019tabfact}), charts (ChartQA~\cite{masry2022chartqa}), sence text (TextVQA~\cite{singh2019towardstextvqa}), and KIE (FUNSD~\cite{jaume2019funsd}, SROIE~\cite{huang2019icdar2019sroie}, POIE~\cite{kuang2023visualpoie}). 
The evaluation metrics used are derived from the official metrics provided. It is important to note that TextVQA is evaluated using the validation set, while the other datasets are evaluated using their respective test sets. As shown in Table \ref{tab:res1},  Marten demonstrates superior performance compared to existing MLLMs, particularly excelling in text-dense and smaller document scenarios. Marten achieve consistently and significantly performence improvements on multiple benchmarks, leading in datasets such as DocVQA, InfoVQA, DeepForm, KLC, WTQ, TabFact, FUNSD, and SROIE, indicating a more comprehensive capability in visual document understanding. Compared to the existing best methods under each benchmark, Marten achieves an average improvement of 1.97\% in document benchmarks, 5.09\% in table benchmarks, and 3.73\% in key information extraction benchmarks. This demonstrates that our alignment strategy aids Marten in better locating the position of visual texts and accurately finding the answers. However, in the chart and sence text benchmarks, Marten's performance is lower than that of InternVL2, which is trained on hundreds of millions of samples. This indicates that Marten still lacks understanding in charts and perception abilities in natural scenes, which will be a focus for future optimization efforts.

\begin{figure}[t]
    \centering
    \includegraphics[width=1.0\linewidth]{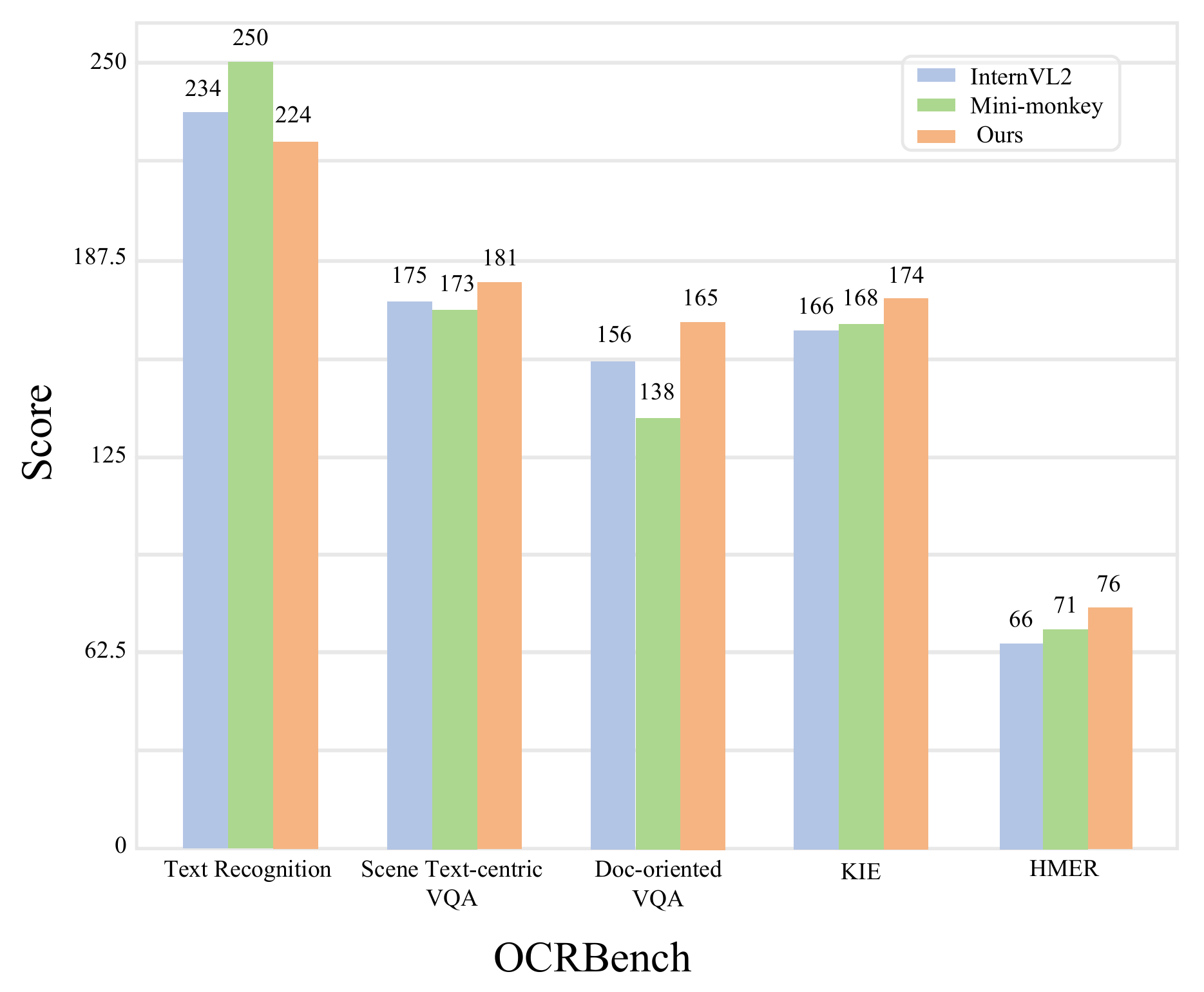}
    \vspace{-2em}
    \caption{Bar chart of scores for each subtask in OCRBench. ``KIE" stands for Key Information Extraction, and ``HMER" stands for Handwritten Mathematical Expression Recognition.}
    \label{fig:ocrbench_show}
    \vspace{-1em}
\end{figure}

\noindent \textbf{OCRBench.} To comprehensively evaluate the performance of Marten, Table \ref{tab:ocrbench} presents a comparison of Marten with existing MLLMs on OCRBench~\cite{liu2023hiddenocrbench}. OCRBench is a recently developed benchmark designed to assess the optical character recognition (OCR) capabilities of MLLMs. It encompasses a wide range of text-related visual tasks, divided into five subtasks: Text Recognition, Scene Text-centric VQA, Doc-oriented VQA, Key Information Extraction (KIE), and Handwritten Mathematical Expression Recognition (HMER). In total, it includes 29 datasets and aims to produce an overall score. Specifically, Marten achieved a score of 820 on OCRBench, which is 26 points higher than InternVL2 and 18 points higher than MiniMonkey, demonstrating Marten's efficient performance across a broad spectrum of text-related visual tasks. Additionally, Figure \ref{fig:ocrbench_show} illustrates Marten's scores compared to recent MLLMs in the five subtasks. It is observed that by employing the VQAMask vision-language alignment method, Marten demonstrates superior performance in both VQA tasks and transcription tasks. It is noteworthy that since the Text Recognition task lacks layout information, our method does not provide effective improvements in this area.

\begin{figure*}[t]
    \centering
    \includegraphics[width=1\linewidth]{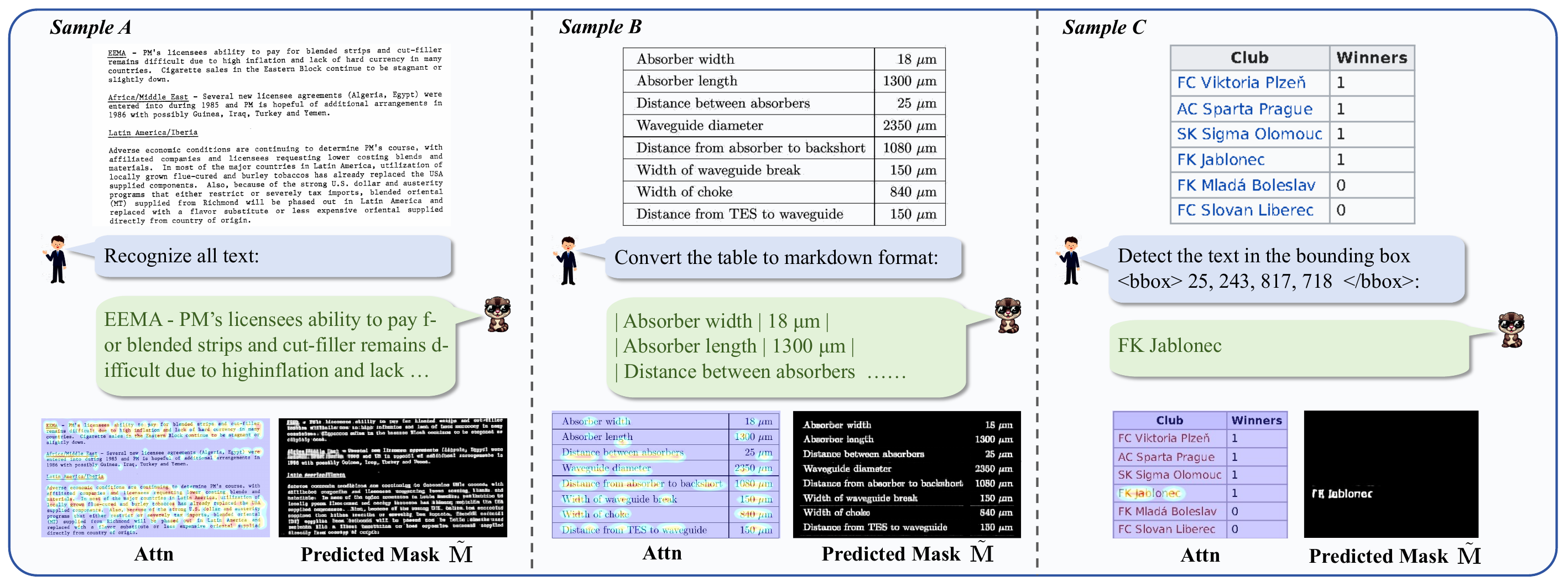}
    \vspace{-1em}
    \caption{Visualization of output results in VQAMask alignment training. We present samples for three different tasks: 1) Sample A represents full-image visual text recognition, 2) Sample B represents Markdown-style transcription, and 3) Sample C represents reading partial text guided by the bounding box.}
    \label{fig:chat}
\end{figure*}

\subsection{Ablation Study}

Extensive ablation experiments are conducted to verify the effectiveness of the module. The results of both the first and second training stages are validated separately. To assess the effectiveness of the MGM, different model combinations are integrated for verification.

\noindent \textbf{Stage 1.} In the first training stage, we compared Marten's performance with and without the MGM, as shown in Table \ref{tab:as_1}. The enhancement of Marten's vision-language alignment capability by MGM is verified through recognition results in both natural and document scenarios. For natural scenes, we use the ICDAR15~\cite{karatzas2015icdar} and TotalText~\cite{ch2017totaltext} datasets. For document scenarios, one thousand images from IIT-CDIP~\cite{iit}, not involved in training, are selected, and PaddleOCR is used to recognize the visual texts, constructing the recognition results for evaluation. Additionally, we extract one thousand latex-formatted tables and equations from DocGenome~\cite{xia2024docgenome}, which are also not used during training, to assess Marten's transcription performance. We discuss the impact of MGM on vision-language alignment under different model combinations. The visual foundation model options include Swin-Transformer~\cite{liu2021swin} and InternViT~\cite{chen2024far}, while the LLM choices are Vicuna1.5~\cite{chiang2023vicuna} and InternLM2~\cite{cai2024internlm2}. Since bounding box information is not included during the training phase, the recognition output is evaluated using Edit Distance. Experimental results indicate that after adding MGM, Marten's average Edit Distance in both natural and document scenarios decreases by 0.06. In transcription tasks, the average edit distance decreases by approximately 0.1, showing a more significant improvement. This indicates that MGM helps align the visual foundation model with the LLM, thereby enhancing the model's ability to recognize and parse visual texts.

In Figure \ref{fig:chat}, we present the visualization output results of Marten during VQAMask alignment training across three tasks: full-image parsing, transcription, and partial text recognition. The binary masks of the outputs for these tasks indicate that Marten generates relatively accurate visual texts, demonstrating the feasibility of the method. Additionally, heatmaps are included to show the regions of the image that Marten focuses on. It is observed that after applying VQAMask alignment training, the LLM's perception of the image is concentrated on areas associated with the QA content, confirming that VQAMask enhances the model's spatial awareness of visual texts.

\begin{table}[t]
    \centering
    \vspace{-1.0em}
    \caption{Ablation study in stage 1. Edit Distance (ED) is used as the evaluation metric. ``BB" refers to the backbone. 
    }
    \vspace{-0.7em}
    \scalebox{0.73}{
    \begin{tabular}{C{5mm}C{17mm}|C{7.5mm}|L{14mm}L{14mm}L{14mm}L{14mm}}
        \toprule
        \multirow{2}{*}{BB} & \multirow{2}{*}{LLM} & \multirow{2}{*}{MGM} & \multicolumn{4}{c}{Edit Distance (ED)} \\
        & & & IC15$\downarrow$ & TT$\downarrow$  & IIT$\downarrow$  & DocG$\downarrow$ \\
        \midrule
        \multirow{2}{*}{Swin} & \multirow{2}{*}{Vicuna1.5} & $\times$ & 0.344 & 0.484 & 0.289 & 0.346 \\
                              &  & $\checkmark$ & \textbf{0.262}\textsubscript{\textbf{(-0.082)}} & \textbf{0.397}\textsubscript{\textbf{(-0.087)}} & \textbf{0.234}\textsubscript{\textbf{(-0.055)}} & \textbf{0.242}\textsubscript{\textbf{(-0.112)}} \\
        \midrule
        \multirow{2}{*}{Swin} & \multirow{2}{*}{InternLM2} & $\times$ & 0.325 & 0.478 & 0.286 & 0.315 \\
        &  & $\checkmark$ & \textbf{0.254}\textsubscript{\textbf{(-0.071)}} & \textbf{0.396}\textsubscript{\textbf{(-0.082)}} & \textbf{0.218}\textsubscript{\textbf{(-0.068)}} & \textbf{0.225}\textsubscript{\textbf{(-0.09)}} \\
        \midrule
        \multirow{2}{*}{ViT} & \multirow{2}{*}{InternLM2} & $\times$ & 0.339 & 0.451 & 0.271 & 0.319 \\
        &  & $\checkmark$ & \textbf{0.278}\textsubscript{\textbf{(-0.061)}} & \textbf{0.381}\textsubscript{\textbf{(-0.07)}} & \textbf{0.193}\textsubscript{\textbf{(-0.078)}} & \textbf{0.211}\textsubscript{\textbf{(-0.108)}} \\
        \bottomrule
    \end{tabular}}
    \label{tab:as_1}
    \vspace{-1em}
\end{table}

\noindent \textbf{Stage 2.} In Table \ref{tab:as_2}, we discuss the improvement in visual document understanding performance brought by MGM under different model combinations. The model configurations remain consistent with those in Table \ref{tab:as_1}. We conduct comparisons on four text-rich image benchmarks, including DocVQA, InfoVQA, ChartQA, and TextVQA. MGM improves performance across all four benchmarks, with a particularly noticeable enhancement in DocVQA. Specifically, in the combination of Swin-Transformer and InternLM2, DocVQA shows an improvement of 4.73\%. However, when the Swin-Transformer is used as the visual foundation model, its performance on InfoVQA is inferior to that of InternViT. This is mainly because InfoVQA consists images with super high aspect ratio, which makes it challenging for Swin-Transformer, without employing a crop strategy, to effectively extract visual texts. 
\begin{table}[t]
    \centering
    \vspace{-1.5em}
    \caption{Ablation study in stage 2. ``BB" refers to the backbone, and ``Val" refers to the validation set. 
    }
    \vspace{-0.7em}
    \scalebox{0.73}{
    \begin{tabular}{C{6mm}C{14mm}|c|L{13mm}L{13mm}L{13mm}L{13mm}}
        \toprule
        BB & LLM & MGM & DocVQA & InfoVQA & ChartQA & TextVQA\textsubscript{Val} \\
        \midrule
        \multirow{2}{*}{Swin} & \multirow{2}{*}{Vicuna1.5} & $\times$ & 78.45 & 43.55 & 69.15 & 71.63 \\
                              &  & $\checkmark$ & \textbf{81.89}\textsubscript{\textbf{(+3.44)}} & \textbf{47.19}\textsubscript{\textbf{(+3.64)}} & \textbf{72.01}\textsubscript{\textbf{(+2.86)}} & \textbf{74.97}\textsubscript{\textbf{(+3.34)}} \\
        \midrule
        \multirow{2}{*}{Swin} & \multirow{2}{*}{InternLM2} & $\times$ & 81.12 & 48.50 & 73.75 & 71.34 \\
        &  & $\checkmark$ & \textbf{85.85}\textsubscript{\textbf{(+4.73)}} & \textbf{52.21}\textsubscript{\textbf{(+3.71)}} & \textbf{76.77}\textsubscript{\textbf{(+3.02)}} & \textbf{74.92}\textsubscript{\textbf{(+3.58)}} \\
        \midrule
        \multirow{2}{*}{ViT} & \multirow{2}{*}{InternLM2} & $\times$ & 89.52 & 71.65 & 79.26 & 71.25 \\
        &  & $\checkmark$ & \textbf{92.01}\textsubscript{\textbf{(+2.49)}} & \textbf{75.21}\textsubscript{\textbf{(+3.56)}} & \textbf{81.72}\textsubscript{\textbf{(+2.46)}} & \textbf{74.38}\textsubscript{\textbf{(+3.13)}} \\
        \bottomrule
    \end{tabular}}
    \label{tab:as_2}
    \vspace{-1em}
\end{table}
\section{Conclusion}
In this study, we introduce a novel visual language alignment method, Visual Question Answering with Mask generation (VQAMask), during the pre-training stage to bridge the gap between visual and language modalities. While keeping LLM weights frozen, VQAMask assists the MLLM in simultaneously conducting VQA-based text parsing and mask generation tasks. This optimization process not only leverages the contextual capabilities of the powerful large language model but also promotes the learning of spatially-aware and semantic-aware feature representations for the image encoder. To achieve this, we establish a comprehensive image-mask generation pipeline, and provide MTMask6M with 6M data. Extensive ablation experiments validate the effectiveness and significance of the proposed VQAMask. Finally, leveraging the proposed VQAMask, we introduce Marten, a training-efficient MLLM tailored for general document-level understanding. In future work, we aim to further explore more fine-grained and robust visual language alignment methods to enhance visual document understanding.

%% file: suppl.tex

\section{More visualizations about VQAMask}
\begin{figure*}[h!]
  \vspace{-2em}
  \centering
  \includegraphics[width=0.9\textwidth]{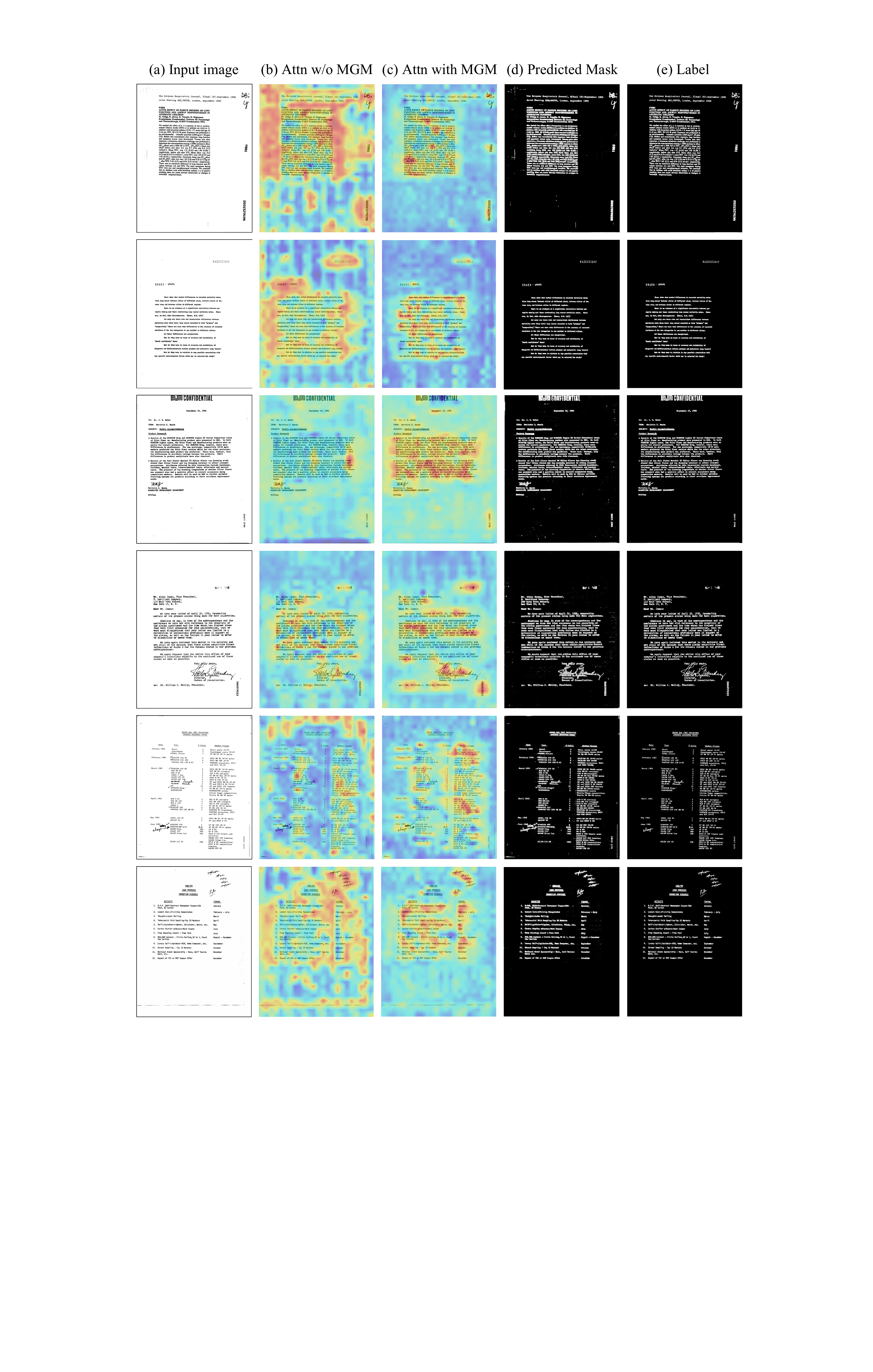}
  \vspace{-1em}
  \caption{Visualizations of some key items in Reading Full
Text task, including (a) Input image (b) Attention without MGM (c) Attention with MGM (d) Prediction Mask and (e) Our generated label. }
  \label{fig2}
\end{figure*}

\begin{figure*}[h!]
  \vspace{-2em}
  \centering
  \includegraphics[width=0.9\textwidth]{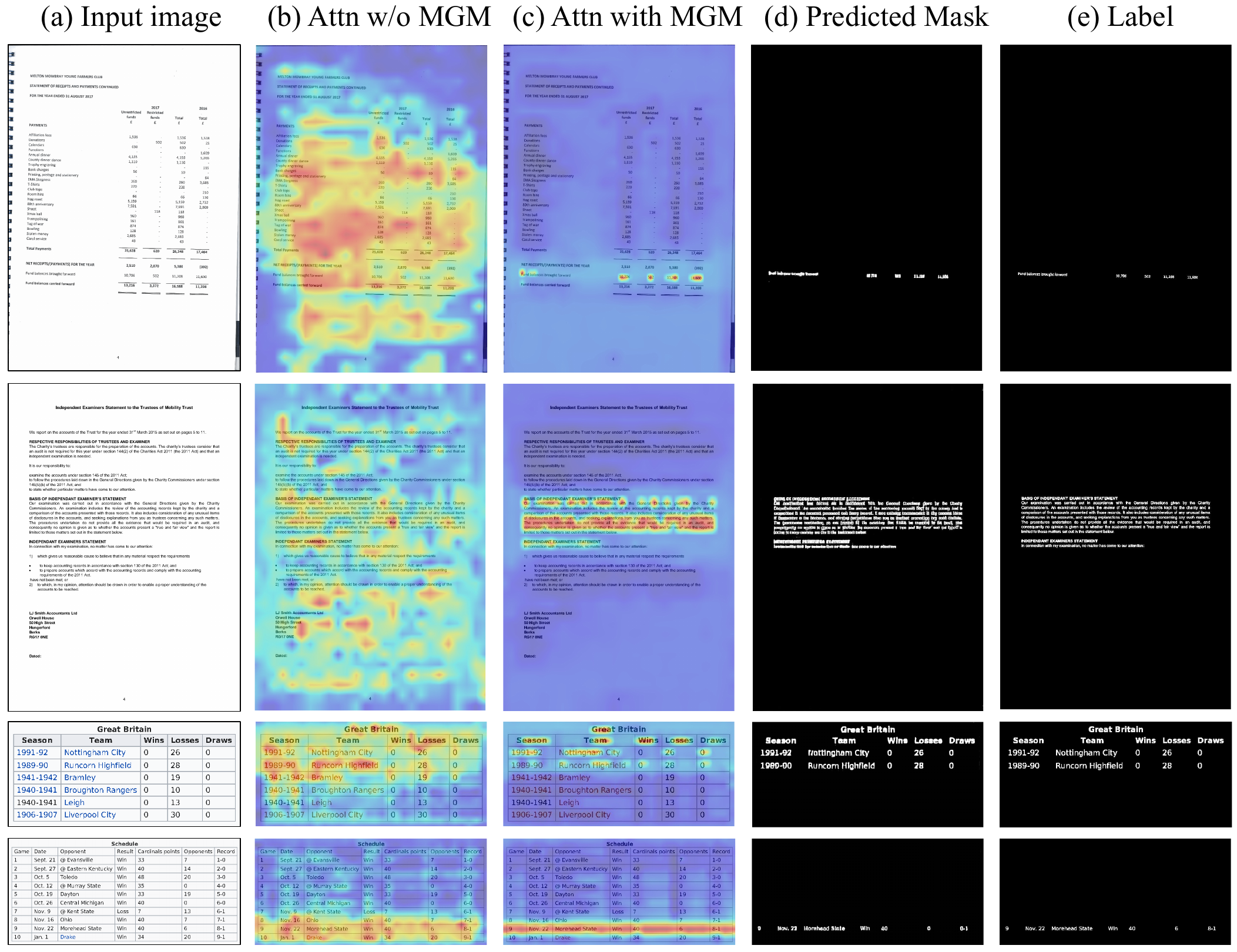}
  \caption{Visualizations of some key items in Reading Partial Text within Localization task, including (a) Input image (b) Attention without MGM (c) Attention with MGM (d) Prediction Mask and (e) Our generated label. }
  \label{fig1}
\end{figure*}

In this section, we show more visualization examples in Figure \ref{fig2} and \ref{fig1}. Each example includes (a) Input image, (b) Attention w/o MGM, (c) Attention with MGM, (d) Prediction Mask, and (e) Our generated label. Specifically, these attention maps in the ``Attention w/o MGM" column (b) are obtained from the version without our proposed mask generation module (MGM). These attention maps in the ``Attention with MGM" column (c) are obtained from the version using our proposed mask generation module (MGM). The ``Predicted Mask" column (d) exhibits the final predicted mask, which delineates all text locations in the document, with spatially-aware supervision by our generated labels (e).  

\noindent \textbf{Example A:} 

Figure \ref{fig2} exhibits the visualizations from the task: \textbf{Reading Full Text.} 
Given an image, the model needs to predict all visual texts sequentially. Specifically, the image, question, and answer are embedded into a question-answer template like: 

\begin{tcolorbox}[colframe=black, colback=white, boxrule=1pt, arc=4pt, width=0.48\textwidth, left=0pt, right=0pt, top=0pt, bottom=0pt]
\noindent \textbf{\textsc{Question:}} \texttt{Recognize all texts.|Convert the image into Markdown format.} 



\noindent \textbf{\textsc{Answer:}} \texttt{BRAND R6D ... SALEM LTS 85.}
\end{tcolorbox}

In this task, our model combines the question and answer to activate the visual text regions of the input image. When comparing the attention maps from the (b) and (c) columns, we observed MGM promotes the alignment between visual tokens and language tokens. In other words, visual tokens corresponding to the visual text regions are further highlighted. The highlighted attentions allow our model to capture more important information for subsequent visual question answering. 

\noindent \textbf{Example B:} 

Figure \ref{fig1} exhibits the examples from the task: \textbf{Reading Partial Text within Localization.} Similarly, the question-answer template is formulated:

\begin{tcolorbox}[colframe=black, colback=white, boxrule=1pt, arc=4pt, width=0.48\textwidth, left=0pt, right=0pt, top=0pt, bottom=0pt]
    \noindent \textbf{\textsc{Question:}} \texttt{Identify the text within the bounding box <bbox> 109, 85, 595, 389  </bbox>.}
    
    \noindent \textbf{\textsc{Answer:}} \texttt{9 Nov.22 Morehead State Win 40 6 8-1.}
\end{tcolorbox}

In this task, the model needs to understand the significance of the number within the \texttt{<bbox>, </bbox>} tags. The number represents a box and its specific location in the image. Only by understanding this can the model accurately predict the text in the box.
Obviously, this task is more challenging. As shown in the second column, the version without our proposed MGM is difficult to find the specific location of the given box. If the location is incorrect, the prediction result will also be wrong. In the version with MGM, with explicit position supervision (presented in the last column), the interaction between language and image can effectively promote the model's understanding of these tokens. As a result, the obtained attention maps are more accurate.

\noindent \textbf{Example C:} 

\begin{figure*}[h!]
  \vspace{-2em}
  \centering
  \includegraphics[width=0.9\textwidth]{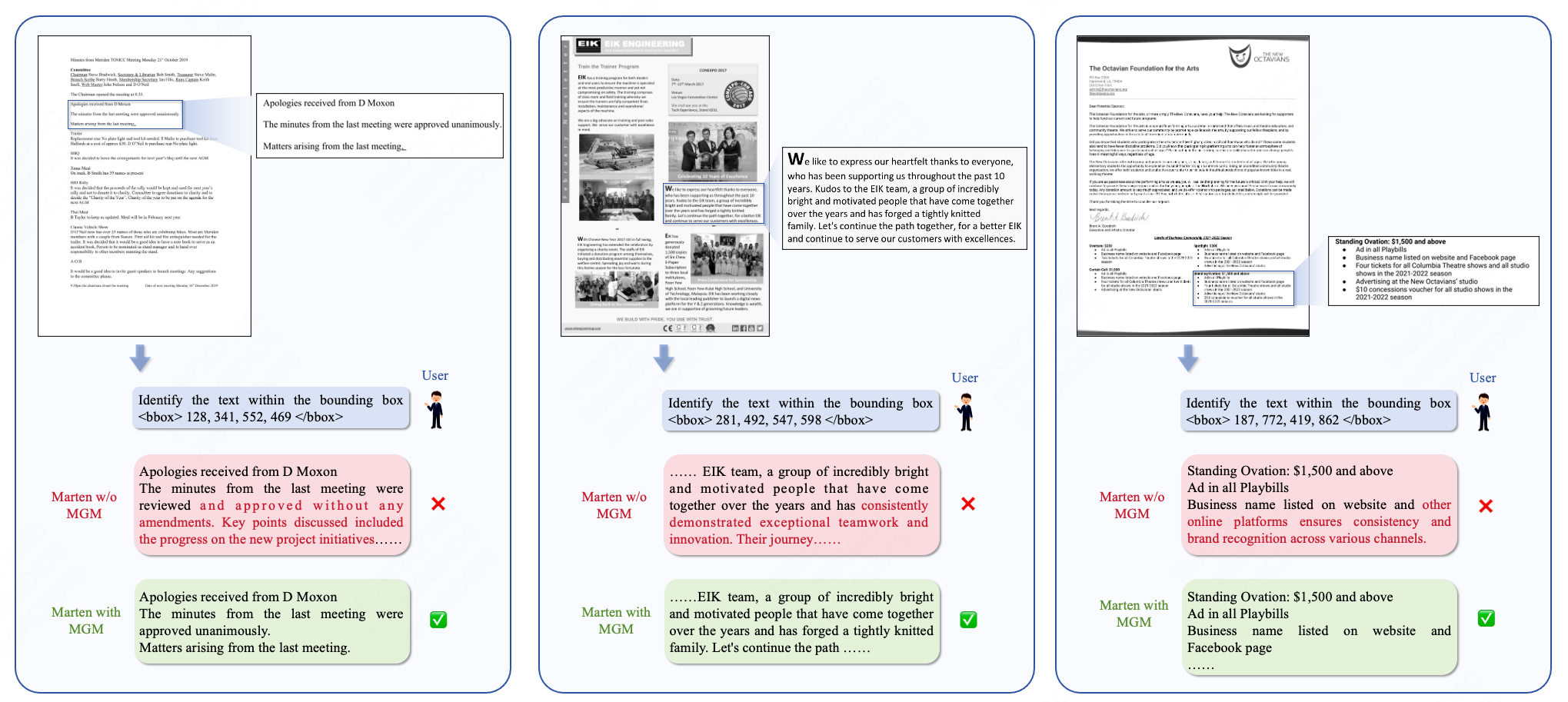}
  \caption{Qualitative comparison results of using and not using MGM.}
  \label{fig3}
\end{figure*}

In Figure \ref{fig3}, we further exhibit the qualitative comparison results of using and not using MGM.
Without spatially-aware supervision, the outputs from the version without MGM may disproportionately rely on the powerful semantic context capabilities of large language models (LLMs) rather than optimizing image features from visual encoders, potentially leading to model hallucinations. As discussed above, our proposed VQAMask optimises two tasks simultaneously: VQA-based text parsing and mask generation. The former allows the model to implicitly align images and text at the semantic level. The latter introduces an additional mask generator (discarded during inference) to explicitly ensure alignment between visual texts within images and their corresponding image regions at a spatially-aware level. Together, they can prevent model hallucinations when parsing visual text and effectively promote spatially-aware feature representation learning.

\clearpage 

\clearpage 

\clearpage

\section{More examples compared to other MLLMs}
\begin{figure*}[htbp]
  \vspace{-2em}
  \centering
  \vspace{-1em}
  \includegraphics[width=0.9\textwidth]{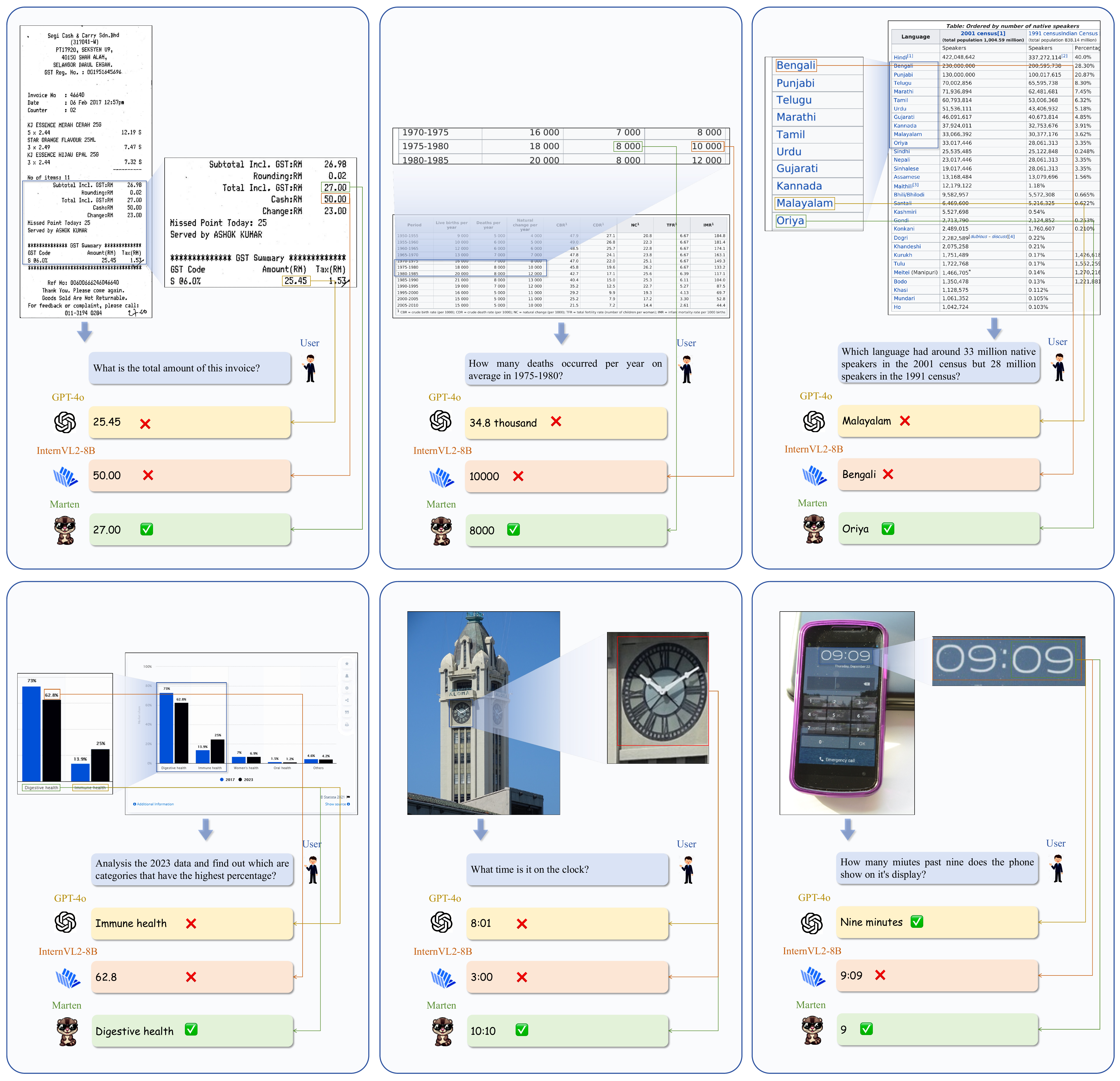}
  \caption{Visualization of Marten’s comparison with GPT-4o, internvl2-8B on VQA tasks.}
  \label{fig4}
\end{figure*}

As shown in Figure \ref{fig4}, we present more qualitative visualization results to demonstrate  Marten’s capabilities in various VQA tasks. Marten analyzes 
the question, \linebreak \newpage \noindent
identifies the key elements in the image relevant to answering the question, and exhibits the impressive localization ability to perceive even minute text within the image.

\newpage

%% file: main.bbl
\begin{thebibliography}{97}
\providecommand{\natexlab}[1]{#1}
\providecommand{\url}[1]{\texttt{#1}}
\expandafter\ifx\csname urlstyle\endcsname\relax
  \providecommand{\doi}[1]{doi: #1}\else
  \providecommand{\doi}{doi: \begingroup \urlstyle{rm}\Url}\fi

\bibitem[Achiam et~al.(2023)Achiam, Adler, Agarwal, Ahmad, Akkaya, Aleman, Almeida, Altenschmidt, Altman, Anadkat, et~al.]{achiam2023gpt}
Josh Achiam, Steven Adler, Sandhini Agarwal, Lama Ahmad, Ilge Akkaya, Florencia~Leoni Aleman, Diogo Almeida, Janko Altenschmidt, Sam Altman, Shyamal Anadkat, et~al.
\newblock Gpt-4 technical report.
\newblock \emph{arXiv preprint arXiv:2303.08774}, 2023.

\bibitem[Bai et~al.(2023)Bai, Bai, Yang, Wang, Tan, Wang, Lin, Zhou, and Zhou]{bai2023qwen}
Jinze Bai, Shuai Bai, Shusheng Yang, Shijie Wang, Sinan Tan, Peng Wang, Junyang Lin, Chang Zhou, and Jingren Zhou.
\newblock Qwen-vl: A frontier large vision-language model with versatile abilities.
\newblock \emph{arXiv preprint arXiv:2308.12966}, 2023.

\bibitem[Biten et~al.(2019)Biten, Tito, Mafla, Gomez, Rusinol, Valveny, Jawahar, and Karatzas]{biten2019scene}
Ali~Furkan Biten, Ruben Tito, Andres Mafla, Lluis Gomez, Mar{\c{c}}al Rusinol, Ernest Valveny, CV Jawahar, and Dimosthenis Karatzas.
\newblock Scene text visual question answering.
\newblock In \emph{Proceedings of the IEEE/CVF international conference on computer vision}, pages 4291--4301, 2019.

\bibitem[Borchmann et~al.(2021)Borchmann, Pietruszka, Stanislawek, Jurkiewicz, Turski, Szyndler, and Grali{\'n}ski]{borchmann2021due}
{\L}ukasz Borchmann, Micha{\l} Pietruszka, Tomasz Stanislawek, Dawid Jurkiewicz, Micha{\l} Turski, Karolina Szyndler, and Filip Grali{\'n}ski.
\newblock Due: End-to-end document understanding benchmark.
\newblock In \emph{Thirty-fifth Conference on Neural Information Processing Systems Datasets and Benchmarks Track (Round 2)}, 2021.

\bibitem[Cai et~al.(2024)Cai, Cao, Chen, Chen, Chen, Chen, Chen, Chen, Chen, and et~al]{cai2024internlm2}
Zheng Cai, Maosong Cao, Haojiong Chen, Kai Chen, Keyu Chen, Xin Chen, Xun Chen, Zehui Chen, Zhi Chen, and et al.
\newblock Internlm2 technical report, 2024.

\bibitem[Caron et~al.(2021)Caron, Touvron, Misra, J{\'e}gou, Mairal, Bojanowski, and Joulin]{DINO}
Mathilde Caron, Hugo Touvron, Ishan Misra, Herv{\'e} J{\'e}gou, Julien Mairal, Piotr Bojanowski, and Armand Joulin.
\newblock Emerging properties in self-supervised vision transformers.
\newblock In \emph{ICCV}, pages 9650--9660, 2021.

\bibitem[Chen et~al.(2019)Chen, Wang, Chen, Zhang, Wang, Li, Zhou, and Wang]{chen2019tabfact}
Wenhu Chen, Hongmin Wang, Jianshu Chen, Yunkai Zhang, Hong Wang, Shiyang Li, Xiyou Zhou, and William~Yang Wang.
\newblock Tabfact: A large-scale dataset for table-based fact verification.
\newblock \emph{arXiv preprint arXiv:1909.02164}, 2019.

\bibitem[Chen et~al.(2024{\natexlab{a}})Chen, Wang, Tian, Ye, Gao, Cui, Tong, Hu, Luo, Ma, et~al.]{chen2024far}
Zhe Chen, Weiyun Wang, Hao Tian, Shenglong Ye, Zhangwei Gao, Erfei Cui, Wenwen Tong, Kongzhi Hu, Jiapeng Luo, Zheng Ma, et~al.
\newblock How far are we to gpt-4v? closing the gap to commercial multimodal models with open-source suites.
\newblock \emph{arXiv preprint arXiv:2404.16821}, 2024{\natexlab{a}}.

\bibitem[Chen et~al.(2024{\natexlab{b}})Chen, Wu, Wang, Su, Chen, Xing, Zhong, Zhang, Zhu, Lu, et~al.]{internvl}
Zhe Chen, Jiannan Wu, Wenhai Wang, Weijie Su, Guo Chen, Sen Xing, Muyan Zhong, Qinglong Zhang, Xizhou Zhu, Lewei Lu, et~al.
\newblock Internvl: Scaling up vision foundation models and aligning for generic visual-linguistic tasks.
\newblock In \emph{Proceedings of the IEEE/CVF Conference on Computer Vision and Pattern Recognition}, pages 24185--24198, 2024{\natexlab{b}}.

\bibitem[Chiang et~al.(2023)Chiang, Li, Lin, Sheng, Wu, Zhang, Zheng, Zhuang, Zhuang, Gonzalez, et~al.]{chiang2023vicuna}
Wei-Lin Chiang, Zhuohan Li, Zi Lin, Ying Sheng, Zhanghao Wu, Hao Zhang, Lianmin Zheng, Siyuan Zhuang, Yonghao Zhuang, Joseph~E Gonzalez, et~al.
\newblock Vicuna: An open-source chatbot impressing gpt-4 with 90\%* chatgpt quality.
\newblock \emph{See https://vicuna. lmsys. org (accessed 14 April 2023)}, 2\penalty0 (3):\penalty0 6, 2023.

\bibitem[Ch'ng and Chan(2017)]{ch2017totaltext}
Chee~Kheng Ch'ng and Chee~Seng Chan.
\newblock Total-text: A comprehensive dataset for scene text detection and recognition.
\newblock In \emph{2017 14th IAPR international conference on document analysis and recognition (ICDAR)}, pages 935--942. IEEE, 2017.

\bibitem[Deng et~al.(2022)Deng, Sun, Lees, Wu, and Yu]{deng2022turl}
Xiang Deng, Huan Sun, Alyssa Lees, You Wu, and Cong Yu.
\newblock Turl: Table understanding through representation learning.
\newblock \emph{ACM SIGMOD Record}, 51\penalty0 (1):\penalty0 33--40, 2022.

\bibitem[Duan et~al.(2024)Duan, Fu, Guo, Jiang, and Wei]{duan2024odm}
Chen Duan, Pei Fu, Shan Guo, Qianyi Jiang, and Xiaoming Wei.
\newblock Odm: A text-image further alignment pre-training approach for scene text detection and spotting.
\newblock In \emph{Proceedings of the IEEE/CVF Conference on Computer Vision and Pattern Recognition}, pages 15587--15597, 2024.

\bibitem[Fan et~al.(2024)Fan, Chen, Liu, Yuan, and Sigal]{fan2024pre}
Wan-Cyuan Fan, Yen-Chun Chen, Mengchen Liu, Lu Yuan, and Leonid Sigal.
\newblock On pre-training of multimodal language models customized for chart understanding.
\newblock \emph{arXiv preprint arXiv:2407.14506}, 2024.

\bibitem[Feng et~al.(2023{\natexlab{a}})Feng, Liu, Liu, Zhou, Li, and Huang]{docpedia}
Hao Feng, Qi Liu, Hao Liu, Wengang Zhou, Houqiang Li, and Can Huang.
\newblock Docpedia: Unleashing the power of large multimodal model in the frequency domain for versatile document understanding.
\newblock \emph{arXiv preprint arXiv:2311.11810}, 2023{\natexlab{a}}.

\bibitem[Feng et~al.(2023{\natexlab{b}})Feng, Liu, Liu, Zhou, Li, and Huang]{feng2023docpedia}
Hao Feng, Qi Liu, Hao Liu, Wengang Zhou, Houqiang Li, and Can Huang.
\newblock Docpedia: Unleashing the power of large multimodal model in the frequency domain for versatile document understanding.
\newblock \emph{arXiv preprint arXiv:2311.11810}, 2023{\natexlab{b}}.

\bibitem[Guan et~al.(2022)Guan, Gu, Lu, Tu, Feng, Wu, and Guan]{guan2022industrial}
Tongkun Guan, Chaochen Gu, Changsheng Lu, Jingzheng Tu, Qi Feng, Kaijie Wu, and Xinping Guan.
\newblock Industrial scene text detection with refined feature-attentive network.
\newblock \emph{IEEE Transactions on Circuits and Systems for Video Technology}, 32\penalty0 (9):\penalty0 6073--6085, 2022.

\bibitem[Guan et~al.(2023{\natexlab{a}})Guan, Gu, Tu, Yang, Feng, Zhao, and Shen]{SIGA}
Tongkun Guan, Chaochen Gu, Jingzheng Tu, Xue Yang, Qi Feng, Yudi Zhao, and Wei Shen.
\newblock Self-supervised implicit glyph attention for text recognition.
\newblock In \emph{IEEE Conf. Comput. Vis. Pattern Recog.}, pages 15285--15294, 2023{\natexlab{a}}.

\bibitem[Guan et~al.(2023{\natexlab{b}})Guan, Shen, Yang, Feng, Jiang, and Yang]{CCD}
Tongkun Guan, Wei Shen, Xue Yang, Qi Feng, Zekun Jiang, and Xiaokang Yang.
\newblock Self-supervised character-to-character distillation for text recognition.
\newblock In \emph{Proceedings of the IEEE/CVF International Conference on Computer Vision}, pages 19473--19484, 2023{\natexlab{b}}.

\bibitem[Guan et~al.(2024)Guan, Lin, Shen, and Yang]{guan2024posformer}
Tongkun Guan, Chengyu Lin, Wei Shen, and Xiaokang Yang.
\newblock Posformer: recognizing complex handwritten mathematical expression with position forest transformer.
\newblock In \emph{European Conference on Computer Vision}, pages 130--147. Springer, 2024.

\bibitem[Guan et~al.(2025{\natexlab{a}})Guan, Shen, and Yang]{guan2025ccdplus}
Tongkun Guan, Wei Shen, and Xiaokang Yang.
\newblock Ccdplus: Towards accurate character to character distillation for text recognition.
\newblock \emph{IEEE Transactions on Pattern Analysis and Machine Intelligence}, 2025{\natexlab{a}}.

\bibitem[Guan et~al.(2025{\natexlab{b}})Guan, Shen, Yang, Wang, and Yang]{guan2025bridging}
Tongkun Guan, Wei Shen, Xue Yang, Xuehui Wang, and Xiaokang Yang.
\newblock Bridging synthetic and real worlds for pre-training scene text detectors.
\newblock In \emph{European Conference on Computer Vision}, pages 428--446. Springer, 2025{\natexlab{b}}.

\bibitem[Gupta et~al.(2016)Gupta, Vedaldi, and Zisserman]{gupta2016synthetic}
Ankush Gupta, Andrea Vedaldi, and Andrew Zisserman.
\newblock Synthetic data for text localisation in natural images.
\newblock In \emph{Proceedings of the IEEE conference on computer vision and pattern recognition}, pages 2315--2324, 2016.

\bibitem[Harley et~al.(2015)Harley, Ufkes, and Derpanis]{harley2015evaluation}
Adam~W Harley, Alex Ufkes, and Konstantinos~G Derpanis.
\newblock Evaluation of deep convolutional nets for document image classification and retrieval.
\newblock In \emph{2015 13th International Conference on Document Analysis and Recognition (ICDAR)}, pages 991--995. IEEE, 2015.

\bibitem[Hong et~al.(2024{\natexlab{a}})Hong, Wang, Ding, Yu, Lv, Wang, Cheng, Huang, Ji, Xue, et~al.]{hong2024cogvlm2}
Wenyi Hong, Weihan Wang, Ming Ding, Wenmeng Yu, Qingsong Lv, Yan Wang, Yean Cheng, Shiyu Huang, Junhui Ji, Zhao Xue, et~al.
\newblock Cogvlm2: Visual language models for image and video understanding.
\newblock \emph{arXiv preprint arXiv:2408.16500}, 2024{\natexlab{a}}.

\bibitem[Hong et~al.(2024{\natexlab{b}})Hong, Wang, Lv, Xu, Yu, Ji, Wang, Wang, Dong, Ding, et~al.]{hong2024cogagent}
Wenyi Hong, Weihan Wang, Qingsong Lv, Jiazheng Xu, Wenmeng Yu, Junhui Ji, Yan Wang, Zihan Wang, Yuxiao Dong, Ming Ding, et~al.
\newblock Cogagent: A visual language model for gui agents.
\newblock In \emph{Proceedings of the IEEE/CVF Conference on Computer Vision and Pattern Recognition}, pages 14281--14290, 2024{\natexlab{b}}.

\bibitem[Hu et~al.(2024{\natexlab{a}})Hu, Xu, Ye, Yan, Zhang, Zhang, Li, Zhang, Jin, Huang, et~al.]{mplug-docowl1.5}
Anwen Hu, Haiyang Xu, Jiabo Ye, Ming Yan, Liang Zhang, Bo Zhang, Chen Li, Ji Zhang, Qin Jin, Fei Huang, et~al.
\newblock mplug-docowl 1.5: Unified structure learning for ocr-free document understanding.
\newblock \emph{arXiv preprint arXiv:2403.12895}, 2024{\natexlab{a}}.

\bibitem[Hu et~al.(2024{\natexlab{b}})Hu, Xu, Zhang, Ye, Yan, Zhang, Jin, Huang, and Zhou]{Mplug-Docowl2}
Anwen Hu, Haiyang Xu, Liang Zhang, Jiabo Ye, Ming Yan, Ji Zhang, Qin Jin, Fei Huang, and Jingren Zhou.
\newblock mplug-docowl2: High-resolution compressing for ocr-free multi-page document understanding.
\newblock \emph{arXiv preprint arXiv:2409.03420}, 2024{\natexlab{b}}.

\bibitem[Huang et~al.(2024)Huang, Liu, Liang, Jin, and Bai]{MiNi-Monkey}
Mingxin Huang, Yuliang Liu, Dingkang Liang, Lianwen Jin, and Xiang Bai.
\newblock Mini-monkey: Alleviate the sawtooth effect by multi-scale adaptive cropping.
\newblock \emph{arXiv preprint arXiv:2408.02034}, 2024.

\bibitem[Huang et~al.(2019)Huang, Chen, He, Bai, Karatzas, Lu, and Jawahar]{huang2019icdar2019sroie}
Zheng Huang, Kai Chen, Jianhua He, Xiang Bai, Dimosthenis Karatzas, Shijian Lu, and CV Jawahar.
\newblock Icdar2019 competition on scanned receipt ocr and information extraction.
\newblock In \emph{2019 International Conference on Document Analysis and Recognition (ICDAR)}, pages 1516--1520. IEEE, 2019.

\bibitem[Jaume et~al.(2019)Jaume, Ekenel, and Thiran]{jaume2019funsd}
Guillaume Jaume, Hazim~Kemal Ekenel, and Jean-Philippe Thiran.
\newblock Funsd: A dataset for form understanding in noisy scanned documents.
\newblock In \emph{2019 International Conference on Document Analysis and Recognition Workshops (ICDARW)}, pages 1--6. IEEE, 2019.

\bibitem[Kafle et~al.(2018)Kafle, Price, Cohen, and Kanan]{kafle2018dvqa}
Kushal Kafle, Brian Price, Scott Cohen, and Christopher Kanan.
\newblock Dvqa: Understanding data visualizations via question answering.
\newblock In \emph{Proceedings of the IEEE conference on computer vision and pattern recognition}, pages 5648--5656, 2018.

\bibitem[Kahou et~al.(2017)Kahou, Michalski, Atkinson, K{\'a}d{\'a}r, Trischler, and Bengio]{kahou2017figureqa}
Samira~Ebrahimi Kahou, Vincent Michalski, Adam Atkinson, {\'A}kos K{\'a}d{\'a}r, Adam Trischler, and Yoshua Bengio.
\newblock Figureqa: An annotated figure dataset for visual reasoning.
\newblock \emph{arXiv preprint arXiv:1710.07300}, 2017.

\bibitem[Karatzas et~al.(2013)Karatzas, Shafait, Uchida, Iwamura, i~Bigorda, Mestre, Mas, Mota, Almazan, and De~Las~Heras]{karatzas2013icdar}
Dimosthenis Karatzas, Faisal Shafait, Seiichi Uchida, Masakazu Iwamura, Lluis~Gomez i Bigorda, Sergi~Robles Mestre, Joan Mas, David~Fernandez Mota, Jon~Almazan Almazan, and Lluis~Pere De~Las~Heras.
\newblock Icdar 2013 robust reading competition.
\newblock In \emph{2013 12th international conference on document analysis and recognition}, pages 1484--1493. IEEE, 2013.

\bibitem[Karatzas et~al.(2015)Karatzas, Gomez-Bigorda, Nicolaou, Ghosh, Bagdanov, Iwamura, Matas, Neumann, Chandrasekhar, Lu, et~al.]{karatzas2015icdar}
Dimosthenis Karatzas, Lluis Gomez-Bigorda, Anguelos Nicolaou, Suman Ghosh, Andrew Bagdanov, Masakazu Iwamura, Jiri Matas, Lukas Neumann, Vijay~Ramaseshan Chandrasekhar, Shijian Lu, et~al.
\newblock Icdar 2015 competition on robust reading.
\newblock In \emph{2015 13th international conference on document analysis and recognition (ICDAR)}, pages 1156--1160. IEEE, 2015.

\bibitem[Kim et~al.(2023)Kim, Lee, Kim, Jung, Park, Kim, Yun, Kil, Lee, and Park]{Cream}
Geewook Kim, Hodong Lee, Daehee Kim, Haeji Jung, Sanghee Park, Yoonsik Kim, Sangdoo Yun, Taeho Kil, Bado Lee, and Seunghyun Park.
\newblock Visually-situated natural language understanding with contrastive reading model and frozen large language models.
\newblock \emph{arXiv preprint arXiv:2305.15080}, 2023.

\bibitem[Kirillov et~al.(2023)Kirillov, Mintun, Ravi, Mao, Rolland, Gustafson, Xiao, Whitehead, Berg, Lo, et~al.]{SAM}
Alexander Kirillov, Eric Mintun, Nikhila Ravi, Hanzi Mao, Chloe Rolland, Laura Gustafson, Tete Xiao, Spencer Whitehead, Alexander~C Berg, Wan-Yen Lo, et~al.
\newblock Segment anything.
\newblock In \emph{ICCV}, pages 4015--4026, 2023.

\bibitem[Krylov et~al.(2021)Krylov, Nosov, and Sovrasov]{krylov2021open}
Ilya Krylov, Sergei Nosov, and Vladislav Sovrasov.
\newblock Open images v5 text annotation and yet another mask text spotter.
\newblock In \emph{Asian Conference on Machine Learning}, pages 379--389. PMLR, 2021.

\bibitem[Kuang et~al.(2023)Kuang, Hua, Liang, Yang, Jiang, Ren, and Bai]{kuang2023visualpoie}
Jianfeng Kuang, Wei Hua, Dingkang Liang, Mingkun Yang, Deqiang Jiang, Bo Ren, and Xiang Bai.
\newblock Visual information extraction in the wild: practical dataset and end-to-end solution.
\newblock In \emph{International Conference on Document Analysis and Recognition}, pages 36--53. Springer, 2023.

\bibitem[Laurençon et~al.(2024)Laurençon, Marafioti, Sanh, and Tronchon]{laurençon2024building}
Hugo Laurençon, Andrés Marafioti, Victor Sanh, and Léo Tronchon.
\newblock Building and better understanding vision-language models: insights and future directions., 2024.

\bibitem[Lee et~al.(2024)Lee, Park, Kim, and Ro]{MOAI}
Byung-Kwan Lee, Beomchan Park, Chae~Won Kim, and Yong~Man Ro.
\newblock Moai: Mixture of all intelligence for large language and vision models.
\newblock \emph{ECCV}, 2024.

\bibitem[Lewis et~al.(2006)Lewis, Agam, Argamon, Frieder, Grossman, and Heard]{iit}
David Lewis, Gady Agam, Shlomo Argamon, Ophir Frieder, David Grossman, and Jefferson Heard.
\newblock Building a test collection for complex document information processing.
\newblock In \emph{Proceedings of the 29th annual international ACM SIGIR conference on Research and development in information retrieval}, pages 665--666, 2006.

\bibitem[Li et~al.(2022)Li, Liu, Guo, Yin, Jiang, Du, Du, Zhu, Lai, Hu, et~al.]{PP-OCRv3}
Chenxia Li, Weiwei Liu, Ruoyu Guo, Xiaoting Yin, Kaitao Jiang, Yongkun Du, Yuning Du, Lingfeng Zhu, Baohua Lai, Xiaoguang Hu, et~al.
\newblock Pp-ocrv3: More attempts for the improvement of ultra lightweight ocr system.
\newblock \emph{arXiv preprint arXiv:2206.03001}, 2022.

\bibitem[Li et~al.(2024)Li, Yang, Liu, Ma, Zhang, Yang, Sun, Liu, and Bai]{Monkey}
Zhang Li, Biao Yang, Qiang Liu, Zhiyin Ma, Shuo Zhang, Jingxu Yang, Yabo Sun, Yuliang Liu, and Xiang Bai.
\newblock Monkey: Image resolution and text label are important things for large multi-modal models.
\newblock In \emph{Proceedings of the IEEE/CVF Conference on Computer Vision and Pattern Recognition}, pages 26763--26773, 2024.

\bibitem[Liao et~al.(2024)Liao, Wang, Li, Wang, Huang, and Jin]{Doclayllm}
Wenhui Liao, Jiapeng Wang, Hongliang Li, Chengyu Wang, Jun Huang, and Lianwen Jin.
\newblock Doclayllm: An efficient and effective multi-modal extension of large language models for text-rich document understanding.
\newblock \emph{arXiv preprint arXiv:2408.15045}, 2024.

\bibitem[Liu et~al.(2024{\natexlab{a}})Liu, Yin, Cao, Jiang, Li, Liu, Jiang, Sun, and Xu]{liu2024hrvda}
Chaohu Liu, Kun Yin, Haoyu Cao, Xinghua Jiang, Xin Li, Yinsong Liu, Deqiang Jiang, Xing Sun, and Linli Xu.
\newblock Hrvda: High-resolution visual document assistant.
\newblock In \emph{Proceedings of the IEEE/CVF Conference on Computer Vision and Pattern Recognition}, pages 15534--15545, 2024{\natexlab{a}}.

\bibitem[Liu et~al.(2024{\natexlab{b}})Liu, Li, Wu, and Lee]{liu2024visual}
Haotian Liu, Chunyuan Li, Qingyang Wu, and Yong~Jae Lee.
\newblock Visual instruction tuning.
\newblock \emph{Advances in neural information processing systems}, 36, 2024{\natexlab{b}}.

\bibitem[Liu et~al.(2023)Liu, Li, Yang, Li, Yin, Liu, Jin, and Bai]{liu2023hiddenocrbench}
Yuliang Liu, Zhang Li, Biao Yang, Chunyuan Li, Xucheng Yin, Cheng-lin Liu, Lianwen Jin, and Xiang Bai.
\newblock On the hidden mystery of ocr in large multimodal models.
\newblock \emph{arXiv preprint arXiv:2305.07895}, 2023.

\bibitem[Liu et~al.(2024{\natexlab{c}})Liu, Yang, Liu, Li, Ma, Zhang, and Bai]{liu2024textmonkey}
Yuliang Liu, Biao Yang, Qiang Liu, Zhang Li, Zhiyin Ma, Shuo Zhang, and Xiang Bai.
\newblock Textmonkey: An ocr-free large multimodal model for understanding document.
\newblock \emph{arXiv preprint arXiv:2403.04473}, 2024{\natexlab{c}}.

\bibitem[Liu et~al.(2021)Liu, Lin, Cao, Hu, Wei, Zhang, Lin, and Guo]{liu2021swin}
Ze Liu, Yutong Lin, Yue Cao, Han Hu, Yixuan Wei, Zheng Zhang, Stephen Lin, and Baining Guo.
\newblock Swin transformer: Hierarchical vision transformer using shifted windows.
\newblock In \emph{Proceedings of the IEEE/CVF international conference on computer vision}, pages 10012--10022, 2021.

\bibitem[Lu et~al.(2024)Lu, Yu, Wang, Ye, Tang, Yang, Wu, Liu, Feng, Wang, et~al.]{LayTextLLM}
Jinghui Lu, Haiyang Yu, Yanjie Wang, Yongjie Ye, Jingqun Tang, Ziwei Yang, Binghong Wu, Qi Liu, Hao Feng, Han Wang, et~al.
\newblock A bounding box is worth one token: Interleaving layout and text in a large language model for document understanding.
\newblock \emph{arXiv preprint arXiv:2407.01976}, 2024.

\bibitem[Lv et~al.(2023)Lv, Huang, Chen, Zhao, Jia, Cui, Ma, Chang, Huang, Wang, et~al.]{lv2023kosmos}
Tengchao Lv, Yupan Huang, Jingye Chen, Yuzhong Zhao, Yilin Jia, Lei Cui, Shuming Ma, Yaoyao Chang, Shaohan Huang, Wenhui Wang, et~al.
\newblock Kosmos-2.5: A multimodal literate model.
\newblock \emph{arXiv preprint arXiv:2309.11419}, 2023.

\bibitem[Lyu et~al.(2022)Lyu, Zhang, Liu, Qiao, Xu, Wu, Yao, Han, Ding, and Wang]{lyu2022maskocr}
Pengyuan Lyu, Chengquan Zhang, Shanshan Liu, Meina Qiao, Yangliu Xu, Liang Wu, Kun Yao, Junyu Han, Errui Ding, and Jingdong Wang.
\newblock Maskocr: Text recognition with masked encoder-decoder pretraining.
\newblock \emph{arXiv preprint arXiv:2206.00311}, 2022.

\bibitem[Mahdavi et~al.(2019)Mahdavi, Zanibbi, Mouchere, Viard-Gaudin, and Garain]{mahdavi2019icdar}
Mahshad Mahdavi, Richard Zanibbi, Harold Mouchere, Christian Viard-Gaudin, and Utpal Garain.
\newblock Icdar 2019 crohme+ tfd: Competition on recognition of handwritten mathematical expressions and typeset formula detection.
\newblock pages 1533--1538, 2019.

\bibitem[Marti and Bunke(2002)]{marti2002iam}
U-V Marti and Horst Bunke.
\newblock The iam-database: an english sentence database for offline handwriting recognition.
\newblock \emph{International journal on document analysis and recognition}, 5:\penalty0 39--46, 2002.

\bibitem[Masry et~al.(2022)Masry, Long, Tan, Joty, and Hoque]{masry2022chartqa}
Ahmed Masry, Do~Xuan Long, Jia~Qing Tan, Shafiq Joty, and Enamul Hoque.
\newblock Chartqa: A benchmark for question answering about charts with visual and logical reasoning.
\newblock \emph{arXiv preprint arXiv:2203.10244}, 2022.

\bibitem[Mathew et~al.(2021)Mathew, Karatzas, and Jawahar]{mathew2021docvqa}
Minesh Mathew, Dimosthenis Karatzas, and CV Jawahar.
\newblock Docvqa: A dataset for vqa on document images.
\newblock In \emph{Proceedings of the IEEE/CVF winter conference on applications of computer vision}, pages 2200--2209, 2021.

\bibitem[Mathew et~al.(2022)Mathew, Bagal, Tito, Karatzas, Valveny, and Jawahar]{mathew2022infographicvqa}
Minesh Mathew, Viraj Bagal, Rub{\`e}n Tito, Dimosthenis Karatzas, Ernest Valveny, and CV Jawahar.
\newblock Infographicvqa.
\newblock In \emph{Proceedings of the IEEE/CVF Winter Conference on Applications of Computer Vision}, pages 1697--1706, 2022.

\bibitem[Methani et~al.(2020)Methani, Ganguly, Khapra, and Kumar]{methani2020plotqa}
Nitesh Methani, Pritha Ganguly, Mitesh~M Khapra, and Pratyush Kumar.
\newblock Plotqa: Reasoning over scientific plots.
\newblock In \emph{Proceedings of the IEEE/CVF Winter Conference on Applications of Computer Vision}, pages 1527--1536, 2020.

\bibitem[Milletari et~al.(2016)Milletari, Navab, and Ahmadi]{milletari2016vdiceloss}
Fausto Milletari, Nassir Navab, and Seyed-Ahmad Ahmadi.
\newblock V-net: Fully convolutional neural networks for volumetric medical image segmentation.
\newblock In \emph{2016 fourth international conference on 3D vision (3DV)}, pages 565--571. Ieee, 2016.

\bibitem[Mishra et~al.(2019)Mishra, Shekhar, Singh, and Chakraborty]{mishra2019ocr}
Anand Mishra, Shashank Shekhar, Ajeet~Kumar Singh, and Anirban Chakraborty.
\newblock Ocr-vqa: Visual question answering by reading text in images.
\newblock In \emph{2019 international conference on document analysis and recognition (ICDAR)}, pages 947--952. IEEE, 2019.

\bibitem[Mouchere et~al.(2014)Mouchere, Viard-Gaudin, Zanibbi, and Garain]{mouchere2014icfhr}
Harold Mouchere, Christian Viard-Gaudin, Richard Zanibbi, and Utpal Garain.
\newblock Icfhr 2014 competition on recognition of on-line handwritten mathematical expressions (crohme 2014).
\newblock pages 791--796, 2014.

\bibitem[Mouch{\`e}re et~al.(2016)Mouch{\`e}re, Viard-Gaudin, Zanibbi, and Garain]{mouchere2016icfhr2016}
Harold Mouch{\`e}re, Christian Viard-Gaudin, Richard Zanibbi, and Utpal Garain.
\newblock Icfhr2016 crohme: Competition on recognition of online handwritten mathematical expressions.
\newblock pages 607--612, 2016.

\bibitem[Park et~al.(2024)Park, Choi, Park, and Han]{parkhierarchical}
Jaeyoo Park, Jin~Young Choi, Jeonghyung Park, and Bohyung Han.
\newblock Hierarchical visual feature aggregation for ocr-free document understanding.
\newblock In \emph{The Thirty-eighth Annual Conference on Neural Information Processing Systems}, 2024.

\bibitem[Pasupat and Liang(2015)]{pasupat2015compositionalwtq}
Panupong Pasupat and Percy Liang.
\newblock Compositional semantic parsing on semi-structured tables.
\newblock \emph{arXiv preprint arXiv:1508.00305}, 2015.

\bibitem[Radford et~al.(2021)Radford, Kim, Hallacy, Ramesh, Goh, Agarwal, Sastry, Askell, Mishkin, Clark, et~al.]{CLIP}
Alec Radford, Jong~Wook Kim, Chris Hallacy, Aditya Ramesh, Gabriel Goh, Sandhini Agarwal, Girish Sastry, Amanda Askell, Pamela Mishkin, Jack Clark, et~al.
\newblock Learning transferable visual models from natural language supervision.
\newblock pages 8748--8763, 2021.

\bibitem[Shao et~al.(2024)Shao, Qian, Xiao, Song, Zong, Wang, Liu, and Li]{VisualCoT}
Hao Shao, Shengju Qian, Han Xiao, Guanglu Song, Zhuofan Zong, Letian Wang, Yu Liu, and Hongsheng Li.
\newblock Visual cot: Unleashing chain-of-thought reasoning in multi-modal language models.
\newblock 2024.

\bibitem[Singh et~al.(2019)Singh, Natarajan, Shah, Jiang, Chen, Batra, Parikh, and Rohrbach]{singh2019towardstextvqa}
Amanpreet Singh, Vivek Natarajan, Meet Shah, Yu Jiang, Xinlei Chen, Dhruv Batra, Devi Parikh, and Marcus Rohrbach.
\newblock Towards vqa models that can read.
\newblock In \emph{Proceedings of the IEEE/CVF conference on computer vision and pattern recognition}, pages 8317--8326, 2019.

\bibitem[Singh et~al.(2021)Singh, Pang, Toh, Huang, Galuba, and Hassner]{singh2021textocr}
Amanpreet Singh, Guan Pang, Mandy Toh, Jing Huang, Wojciech Galuba, and Tal Hassner.
\newblock Textocr: Towards large-scale end-to-end reasoning for arbitrary-shaped scene text.
\newblock In \emph{Proceedings of the IEEE/CVF conference on computer vision and pattern recognition}, pages 8802--8812, 2021.

\bibitem[Song et~al.(2022)Song, Wan, Yang, Tang, Cheng, Bai, and Yao]{VLPT}
Sibo Song, Jianqiang Wan, Zhibo Yang, Jun Tang, Wenqing Cheng, Xiang Bai, and Cong Yao.
\newblock Vision-language pre-training for boosting scene text detectors.
\newblock In \emph{CVPR}, pages 15681--15691, 2022.

\bibitem[Stanis{\l}awek et~al.(2021)Stanis{\l}awek, Grali{\'n}ski, Wr{\'o}blewska, Lipi{\'n}ski, Kaliska, Rosalska, Topolski, and Biecek]{stanislawek2021kleister}
Tomasz Stanis{\l}awek, Filip Grali{\'n}ski, Anna Wr{\'o}blewska, Dawid Lipi{\'n}ski, Agnieszka Kaliska, Paulina Rosalska, Bartosz Topolski, and Przemys{\l}aw Biecek.
\newblock Kleister: key information extraction datasets involving long documents with complex layouts.
\newblock In \emph{International Conference on Document Analysis and Recognition}, pages 564--579. Springer, 2021.

\bibitem[Svetlichnaya(2020)]{svetlichnaya2020deepform}
S Svetlichnaya.
\newblock Deepform: Understand structured documents at scale.
\newblock 2020.

\bibitem[Tanaka et~al.(2021)Tanaka, Nishida, and Yoshida]{tanaka2021visualmrc}
Ryota Tanaka, Kyosuke Nishida, and Sen Yoshida.
\newblock Visualmrc: Machine reading comprehension on document images.
\newblock In \emph{Proceedings of the AAAI Conference on Artificial Intelligence}, pages 13878--13888, 2021.

\bibitem[Tanaka et~al.(2024)Tanaka, Iki, Nishida, Saito, and Suzuki]{Instructdoc}
Ryota Tanaka, Taichi Iki, Kyosuke Nishida, Kuniko Saito, and Jun Suzuki.
\newblock Instructdoc: A dataset for zero-shot generalization of visual document understanding with instructions.
\newblock In \emph{AAAI}, pages 19071--19079, 2024.

\bibitem[Turski et~al.(2023)Turski, Stanis{\l}awek, Kaczmarek, Dyda, and Grali{\'n}ski]{turski2023ccpdf}
Micha{\l} Turski, Tomasz Stanis{\l}awek, Karol Kaczmarek, Pawe{\l} Dyda, and Filip Grali{\'n}ski.
\newblock Ccpdf: Building a high quality corpus for visually rich documents from web crawl data.
\newblock In \emph{International Conference on Document Analysis and Recognition}, pages 348--365. Springer, 2023.

\bibitem[Wan et~al.(2021)Wan, Ji, and Shen]{SKTM}
Qi Wan, Haoqin Ji, and Linlin Shen.
\newblock Self-attention based text knowledge mining for text detection.
\newblock In \emph{CVPR}, pages 5983--5992, 2021.

\bibitem[Wang et~al.(2024{\natexlab{a}})Wang, Gu, Liang, Xu, Zhang, Shi, and He]{wang2024unimernet}
Bin Wang, Zhuangcheng Gu, Guang Liang, Chao Xu, Bo Zhang, Botian Shi, and Conghui He.
\newblock Unimernet: A universal network for real-world mathematical expression recognition.
\newblock \emph{arXiv preprint arXiv:2404.15254}, 2024{\natexlab{a}}.

\bibitem[Wang et~al.(2024{\natexlab{b}})Wang, Raman, Sibue, Ma, Babkin, Kaur, Pei, Nourbakhsh, and Liu]{DocLLM}
Dongsheng Wang, Natraj Raman, Mathieu Sibue, Zhiqiang Ma, Petr Babkin, Simerjot Kaur, Yulong Pei, Armineh Nourbakhsh, and Xiaomo Liu.
\newblock Docllm: A layout-aware generative language model for multimodal document understanding.
\newblock page 8529–8548, 2024{\natexlab{b}}.

\bibitem[Wang et~al.(2020)Wang, Liu, Shen, Ng, Luo, Jin, Chan, Hengel, and Wang]{wang2020general}
Xinyu Wang, Yuliang Liu, Chunhua Shen, Chun~Chet Ng, Canjie Luo, Lianwen Jin, Chee~Seng Chan, Anton van~den Hengel, and Liangwei Wang.
\newblock On the general value of evidence, and bilingual scene-text visual question answering.
\newblock In \emph{Proceedings of the IEEE/CVF Conference on Computer Vision and Pattern Recognition}, pages 10126--10135, 2020.

\bibitem[Wei et~al.(2025)Wei, Kong, Chen, Zhao, Ge, Yang, Sun, Han, and Zhang]{vary}
Haoran Wei, Lingyu Kong, Jinyue Chen, Liang Zhao, Zheng Ge, Jinrong Yang, Jianjian Sun, Chunrui Han, and Xiangyu Zhang.
\newblock Vary: Scaling up the vision vocabulary for large vision-language model.
\newblock In \emph{ECCV}, pages 408--424. Springer, 2025.

\bibitem[Wu et~al.(2024)Wu, Yang, Chai, Zhang, Liu, Du, Liang, Shu, Cheng, Sun, et~al.]{wu2024tablebench}
Xianjie Wu, Jian Yang, Linzheng Chai, Ge Zhang, Jiaheng Liu, Xinrun Du, Di Liang, Daixin Shu, Xianfu Cheng, Tianzhen Sun, et~al.
\newblock Tablebench: A comprehensive and complex benchmark for table question answering.
\newblock \emph{arXiv preprint arXiv:2408.09174}, 2024.

\bibitem[Xia et~al.(2024)Xia, Mao, Yan, Zhou, Zhang, Peng, Pi, Fu, Wu, Ye, et~al.]{xia2024docgenome}
Renqiu Xia, Song Mao, Xiangchao Yan, Hongbin Zhou, Bo Zhang, Haoyang Peng, Jiahao Pi, Daocheng Fu, Wenjie Wu, Hancheng Ye, et~al.
\newblock Docgenome: An open large-scale scientific document benchmark for training and testing multi-modal large language models.
\newblock \emph{arXiv preprint arXiv:2406.11633}, 2024.

\bibitem[Xie et~al.(2024)Xie, Yin, Yan, Liu, Ding, Liao, Liu, Chen, and Bai]{xie2024wukong}
Xudong Xie, Liang Yin, Hao Yan, Yang Liu, Jing Ding, Minghui Liao, Yuliang Liu, Wei Chen, and Xiang Bai.
\newblock Wukong: A large multimodal model for efficient long pdf reading with end-to-end sparse sampling.
\newblock \emph{arXiv preprint arXiv:2410.05970}, 2024.

\bibitem[Xue et~al.(2022)Xue, Zhang, Hao, Lu, Torr, and Bai]{oCLIP}
Chuhui Xue, Wenqing Zhang, Yu Hao, Shijian Lu, Philip~HS Torr, and Song Bai.
\newblock Language matters: A weakly supervised vision-language pre-training approach for scene text detection and spotting.
\newblock In \emph{ECCV}, pages 284--302. Springer, 2022.

\bibitem[Yang et~al.(2021)Yang, Lu, Wang, Yin, Florencio, Wang, Zhang, Zhang, and Luo]{yang2021tap}
Zhengyuan Yang, Yijuan Lu, Jianfeng Wang, Xi Yin, Dinei Florencio, Lijuan Wang, Cha Zhang, Lei Zhang, and Jiebo Luo.
\newblock Tap: Text-aware pre-training for text-vqa and text-caption.
\newblock In \emph{Proceedings of the IEEE/CVF conference on computer vision and pattern recognition}, pages 8751--8761, 2021.

\bibitem[Ye et~al.(2023{\natexlab{a}})Ye, Hu, Xu, Ye, Yan, Dan, Zhao, Xu, Li, Tian, et~al.]{ye2023mplug}
Jiabo Ye, Anwen Hu, Haiyang Xu, Qinghao Ye, Ming Yan, Yuhao Dan, Chenlin Zhao, Guohai Xu, Chenliang Li, Junfeng Tian, et~al.
\newblock mplug-docowl: Modularized multimodal large language model for document understanding.
\newblock \emph{arXiv preprint arXiv:2307.02499}, 2023{\natexlab{a}}.

\bibitem[Ye et~al.(2023{\natexlab{b}})Ye, Hu, Xu, Ye, Yan, Xu, Li, Tian, Qian, Zhang, et~al.]{UReader}
Jiabo Ye, Anwen Hu, Haiyang Xu, Qinghao Ye, Ming Yan, Guohai Xu, Chenliang Li, Junfeng Tian, Qi Qian, Ji Zhang, et~al.
\newblock Ureader: Universal ocr-free visually-situated language understanding with multimodal large language model.
\newblock \emph{arXiv preprint arXiv:2310.05126}, 2023{\natexlab{b}}.

\bibitem[Yu et~al.(2023)Yu, Liu, Hua, Jiang, Ren, and Bai]{TCM}
Wenwen Yu, Yuliang Liu, Wei Hua, Deqiang Jiang, Bo Ren, and Xiang Bai.
\newblock Turning a clip model into a scene text detector.
\newblock In \emph{CVPR}, pages 6978--6988, 2023.

\bibitem[Yu et~al.(2024{\natexlab{a}})Yu, Liu, Zhu, Cao, Sun, and Bai]{FastTCM}
Wenwen Yu, Yuliang Liu, Xingkui Zhu, Haoyu Cao, Xing Sun, and Xiang Bai.
\newblock Turning a clip model into a scene text spotter.
\newblock \emph{IEEE TPAMI}, 2024{\natexlab{a}}.

\bibitem[Yu et~al.(2024{\natexlab{b}})Yu, Liao, Wu, Liao, Zheng, and Zeng]{yu2024texthawk}
Ya-Qi Yu, Minghui Liao, Jihao Wu, Yongxin Liao, Xiaoyu Zheng, and Wei Zeng.
\newblock Texthawk: Exploring efficient fine-grained perception of multimodal large language models.
\newblock \emph{arXiv preprint arXiv:2404.09204}, 2024{\natexlab{b}}.

\bibitem[Yu et~al.(2024{\natexlab{c}})Yu, Liao, Zhang, and Wu]{yu2024texthawk2}
Ya-Qi Yu, Minghui Liao, Jiwen Zhang, and Jihao Wu.
\newblock Texthawk2: A large vision-language model excels in bilingual ocr and grounding with 16x fewer tokens.
\newblock \emph{arXiv preprint arXiv:2410.05261}, 2024{\natexlab{c}}.

\bibitem[Zhai et~al.(2023)Zhai, Mustafa, Kolesnikov, and Beyer]{SigLIP}
Xiaohua Zhai, Basil Mustafa, Alexander Kolesnikov, and Lucas Beyer.
\newblock Sigmoid loss for language image pre-training.
\newblock In \emph{ICCV}, pages 11975--11986, 2023.

\bibitem[Zhang et~al.(2024{\natexlab{a}})Zhang, Gao, Gan, Dufter, Wenzel, Huang, Shah, Du, Zhang, Li, et~al.]{zhang2024mm1}
Haotian Zhang, Mingfei Gao, Zhe Gan, Philipp Dufter, Nina Wenzel, Forrest Huang, Dhruti Shah, Xianzhi Du, Bowen Zhang, Yanghao Li, et~al.
\newblock Mm1. 5: Methods, analysis \& insights from multimodal llm fine-tuning.
\newblock \emph{arXiv preprint arXiv:2409.20566}, 2024{\natexlab{a}}.

\bibitem[Zhang et~al.(2024{\natexlab{b}})Zhang, Yang, Lai, Xie, and Jin]{zhang2024dockylin}
Jiaxin Zhang, Wentao Yang, Songxuan Lai, Zecheng Xie, and Lianwen Jin.
\newblock Dockylin: A large multimodal model for visual document understanding with efficient visual slimming.
\newblock \emph{arXiv preprint arXiv:2406.19101}, 2024{\natexlab{b}}.

\bibitem[Zhang et~al.(2024{\natexlab{c}})Zhang, Lyu, Shao, Chen, Guan, and Nie]{zhang2024token}
Renshan Zhang, Yibo Lyu, Rui Shao, Gongwei Chen, Weili Guan, and Liqiang Nie.
\newblock Token-level correlation-guided compression for efficient multimodal document understanding.
\newblock \emph{arXiv preprint arXiv:2407.14439}, 2024{\natexlab{c}}.

\bibitem[Zhong et~al.(2020)Zhong, ShafieiBavani, and Jimeno~Yepes]{zhong2020image}
Xu Zhong, Elaheh ShafieiBavani, and Antonio Jimeno~Yepes.
\newblock Image-based table recognition: data, model, and evaluation.
\newblock In \emph{European conference on computer vision}, pages 564--580. Springer, 2020.

\bibitem[Zhu et~al.(2023)Zhu, Chen, Shen, Li, and Elhoseiny]{minigpt}
Deyao Zhu, Jun Chen, Xiaoqian Shen, Xiang Li, and Mohamed Elhoseiny.
\newblock Minigpt-4: Enhancing vision-language understanding with advanced large language models.
\newblock \emph{arXiv preprint arXiv:2304.10592}, 2023.

\end{thebibliography}
